\documentclass[final,5p,times,twocolumn]{elsarticle}

\usepackage{booktabs}
\usepackage{multirow}
\usepackage{hyperref}
\usepackage{xurl}
\usepackage{xcolor}
\usepackage{bbding}
\usepackage{makecell}
\usepackage{graphicx}
\usepackage{subcaption}
\usepackage{adjustbox}
\usepackage{tabularx}
\usepackage{array}
\usepackage{graphicx}
\usepackage{amsmath, amssymb}
\usepackage[ruled,vlined]{algorithm2e}
\usepackage{multirow}
\usepackage{booktabs}
\usepackage{threeparttable}
\usepackage{longtable}
\usepackage{verbatim}
\usepackage[table]{xcolor}

\usepackage[utf8]{inputenc}
\usepackage{amssymb}
\usepackage{amsmath}
\usepackage[T1]{fontenc}
\hypersetup{hidelinks}

\begin{document}

\begin{frontmatter}

\title{SurgPace: Complexity-Aware Progressive Distillation for Surgical Multimodal Large Language Models} %

\author[1]{Zhicheng He} %
\ead{herichard88@gmail.com}

\author[2]{Xiaohan Xing} %

\author[3]{Hongqiu Wang} %

\author[4]{Kun Yuan} %

\author[1]{Guankun Wang} %

\author[5]{Yanheng Li}

\author[5]{Haoxuan Che}

\author[4]{Nassir Navab} %

\author[1]{Hongliang Ren} %

\author[1,4]{Long Bai\corref{cor1}} %
\ead{b.long@link.cuhk.edu.hk}

\affiliation[1]{organization={Department of Electronic Engineering, The Chinese University of Hong Kong},
    city={Hong Kong SAR},
    country={China}}

\affiliation[2]{organization={Department of Radiation Oncology, Stanford University},
    city={Palo Alto},
    country={USA}}
    
\affiliation[3]{organization={Systems Hub, The Hong Kong University of Science and Technology (Guangzhou)},
    city={Guangzhou},
    country={China}}

\affiliation[4]{organization={Chair of Computer Aided Medical Procedures, Technical University of Munich},
    city={Munich},
    country={Germany}}

\affiliation[5]{organization={XGEN Labs},
    city={Hong Kong SAR},
    country={China}}

\cortext[cor1]{Corresponding Author.}

\begin{abstract}
Multimodal large language models (MLLMs) have shown strong potential in robotic surgery for question answering, scene understanding, and grounded assistance, but their high computational cost limits real clinical deployment. Knowledge distillation provides a practical way to build compact surgical MLLMs, yet direct distillation is not well suited to surgical data because surgical samples vary greatly in difficulty, visual ambiguity, and required clinical context, making uniform teacher-student transfer unstable. In addition, the role of supervision changes during training: early stages require stronger teacher guidance for multimodal alignment, while later stages rely more on task supervision for effective optimization. To address these challenges, we propose SurgPace, a complexity-aware continuous curriculum framework for progressive distillation of surgical vision-language models. SurgPace estimates sample complexity using predictive confidence and Lorentz geometric consistency, and uses the fused score to guide a continuous easy-to-hard training process. We further introduce hierarchical representation alignment, which aligns the student with the teacher at the feature, logit, and probability levels, while gradually shifting the training focus from teacher imitation to task supervision. This design separates sample-level curriculum from objective-level scheduling, leading to more stable and effective distillation. We evaluate SurgPace on Surg-396K, a benchmark built from four public surgical datasets and covering five dialogue paradigms. SurgPace consistently outperforms strong MLLM distillation baselines, demonstrating its effectiveness for building compact, memory-efficient surgical MLLMs  with faster decoding under controlled benchmarks.
Code is available at \url{https://github.com/hezhicheng2002/SurgPace}.
\end{abstract}


\end{frontmatter}

\section{Introduction}
\label{sec:introduction}
Modern surgical care increasingly relies on digital imaging and video streams to support clinical decision-making~\cite{rueckert2026comparative,maier2017surgical}. During minimally invasive procedures, large volumes of visual data are generated, including endoscopic videos, surgical tool trajectories, and intraoperative imaging. These data encode rich information about anatomical structures, surgical actions, and instrument-tissue interactions. Surgeons must continuously interpret these visual signals to infer the current surgical state and make timely decisions~\cite{yang2026large,bai2023surgical}. As surgical procedures become more complex, the volume and heterogeneity of visual data increase rapidly, which further elevates the cognitive burden on clinicians. In recent years, multimodal large language models (MLLMs) have emerged as a promising paradigm for interpreting such complex medical data~\cite{li2023llava,jiang2025hulu,zeng2025surgvlm}. By integrating visual perception with language understanding, these models can analyze surgical scenes while interacting with clinicians through natural language. This capability supports applications such as surgical question answering, workflow understanding, procedure summarization, and clinical dialogue assistance. As a result, MLLMs are increasingly regarded as a potential foundation for intelligent surgical assistants that can observe procedures, interpret context, and communicate clinically relevant information during operations~\cite{schmidgall2024general,wang2024surgical,hao2025enhancing,hao2025surgery}.

\begin{figure*}[t]
    \centering
    \includegraphics[width=\textwidth]{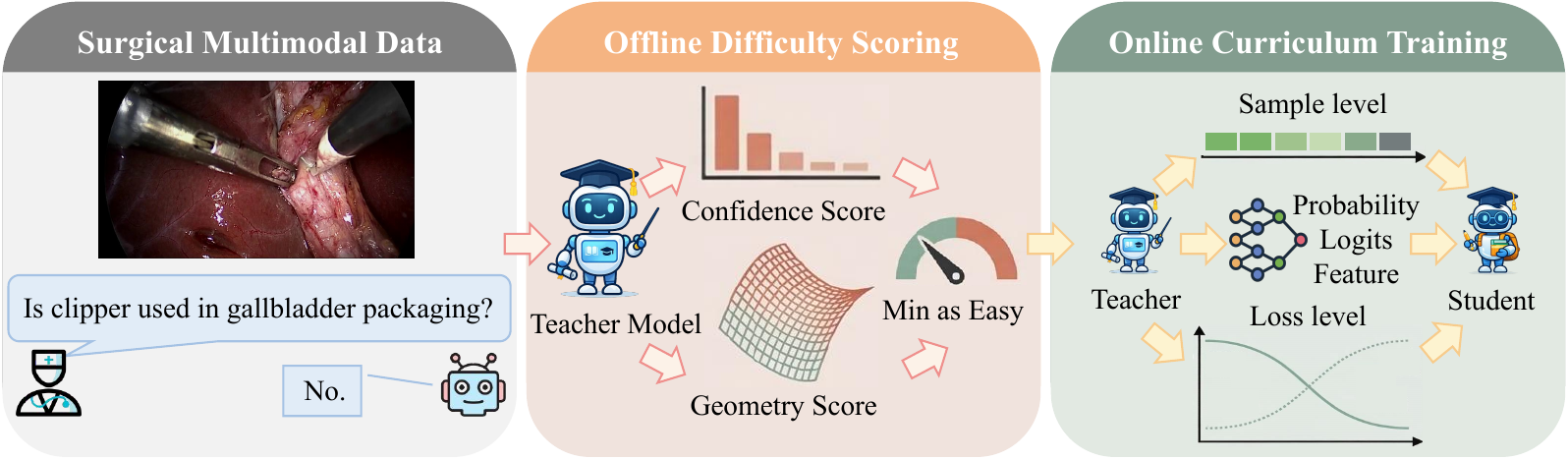}
    \caption{
    Overview of the proposed SurgPace framework.
    Given surgical multimodal data, a teacher model first estimates sample complexity offline using complementary confidence and geometric consistency signals, which are permissively fused so that lower scores indicate easier samples.
    The resulting complexity estimates are then used to guide online curriculum training, where easier samples are emphasized earlier and harder samples are gradually incorporated as training progresses.
    During this process, the student model learns from the teacher through hierarchical representation alignment at the probability, logit, and feature levels, while a progressive objective schedule adjusts the balance of supervision over time.
    Together, these components enable stable and efficient compression of surgical MLLMs into a lightweight student model with reduced computational demands.
    }
    \label{fig:visual_abstract}
\end{figure*}

Despite this promise, the practical deployment of MLLMs in surgical environments remains challenging~\cite{perez2026sureon,he2025parameter,dong2026llama32}. Modern MLLMs typically rely on large vision encoders and deep language backbones, which incur substantial computational overhead. These architectures require considerable memory, high-end GPUs, and long inference time~\cite{tripathi2024efficient}. Such demands conflict with the constraints of real clinical systems, where latency, hardware resources, and system reliability are strictly limited. In operating rooms and robotic surgery platforms, intelligent systems must respond rapidly and reliably, as surgeons often require immediate feedback during critical procedural steps~\cite{zeng2026surg,low2026surgraw}. However, MLLMs are usually developed in research settings with abundant computational resources. Their translation into real clinical workflows therefore remains nontrivial~\cite{ke2026real}. This gap between model capability and deployment feasibility has become a major barrier to practical surgical multimodal intelligence in real clinical settings.

Transferring multimodal capability from a high-capacity teacher model to a resource-efficient student model remains a fundamental challenge for surgical MLLM systems. Knowledge distillation provides a practical route for compressing MLLMs into compact models with lower computational demands~\cite{fang2026knowledge,gu2024minillm,Zhang2024dual}. However, directly applying standard distillation strategies to surgical multimodal tasks often leads to unstable optimization and suboptimal transfer. Several aspects of surgical multimodal learning make this transfer difficult. First, surgical samples differ widely in visual clarity, question complexity, and required clinical context~\cite{Ong2024SurgeryLLM,zhang2025SurGo}. Treating all samples in the same way can cause hard cases to dominate early training and reduce training stability. Second, surgical tasks require joint reasoning over visual appearance, instruments, anatomy, spatial relations, actions, and procedural context~\cite{Yuan2024HecVL}. Distilling only final responses can overlook intermediate cues that support reliable surgical understanding. Third, the useful supervision signal changes during training. Early stages benefit more from teacher-guided supervision for visual and multimodal alignment, while later stages need stronger task supervision to refine final answers. A fixed objective cannot model this change well. These issues limit reliable knowledge transfer in resource-constrained surgical systems, such as operating rooms and robotic platforms.

To address these challenges, we propose SurgPace, a complexity-aware continuous curriculum distillation framework for building compact surgical MLLMs. 
SurgPace trains the student with difficulty-aware sample pacing and progressive multi-level alignment. 
At the sample level, SurgPace estimates the complexity of each multimodal instance using predictive confidence and Lorentz geometric consistency from the teacher. The two measures are fused into one complexity score, so that lower-complexity samples receive larger weights in early training, and this bias is gradually reduced as training proceeds.
At the objective level, SurgPace uses hierarchical representation alignment to operationalize structured knowledge transfer. The student is aligned with the teacher at probability, logit, and hidden-representation levels, while the balance between distillation loss and task loss changes over time. Early training focuses more on teacher-guided transfer, while later training gives more weight to ground-truth supervision. This design separates sample-level curriculum from objective-level scheduling and gives the student a stable learning path. Figure~\ref{fig:visual_abstract} provides an overview of SurgPace and its training workflow.
Specifically, the contributions of this work are summarized as follows:
\begin{itemize}
\item We formulate SurgPace around two coordinated axes: sample-level pacing controls the emphasis on training examples, and objective-level scheduling controls the balance between teacher and task supervision.
\item We combine teacher predictive confidence and Lorentz trajectory dispersion into a complexity score that guides continuous easy-to-hard weighting within hierarchical feature-, logit-, and probability-level alignment.
\item We validate the SurgPace framework on the Surg-396K benchmark, which is built from four public surgical datasets and covers five dialogue paradigms. Extensive experiments and ablations demonstrate strong performance compared to state-of-the-art methods.
\end{itemize}

\section{Related Work}
\label{sec:related_work}
\subsection{Multimodal Large Language Models in Surgical Scenarios}
Multimodal large language models (MLLMs) are shifting surgical AI from single-task, procedure-specific predictors toward interactive assistants that jointly model visual perception, domain knowledge, and natural-language reasoning~\cite{rau2025systematic}. Early surgical MLLM systems primarily focused on tool-use orchestration: VS-Assistant~\cite{Chen2024VSAssistant} translates surgeon intent into structured sequences of algorithmic calls via function-calling tuning, and SurgFC~\cite{Chen2024SurgFC} further unifies intention understanding with downstream vision modules into a single surgical interface. However, such orchestration-centric designs remain constrained by limited visual representation and domain grounding. To address this, GP-VLS~\cite{Schmidgall2024GPVLS} injects surgical textbooks and medical knowledge into a general-purpose surgical vision language model (VLM), enabling broader procedural competence, yet it remains largely image-centric and insufficient for fine-grained temporal reasoning in long operations.

To overcome these limitations, recent work has pursued richer visual grounding and broader multimodal dialogue capabilities. LLaVA-Surg~\cite{Li2024LLaVASurg} introduces large-scale structured surgical video instruction tuning to support multi-turn dialogue across diverse procedures, while EndoChat~\cite{wang2026endochat} advances grounded surgical dialogue through explicit multi-scale visual tokenization for endoscopic scenes. In parallel, SurgVLP~\cite{yuan2025learning} improves surgical language-vision alignment via contrastive pretraining on lecture video-text pairs but still requires task-specific adaptation for interactive intraoperative question-answering (QA). Building on these foundations, SurgVidLM~\cite{Wang2025SurgVidLM} and SurgLLM~\cite{Chen2025SurgLLM} further strengthen surgical multimodal understanding through enhanced visual modeling and context-aware pretraining strategies. Despite these advances, intraoperative deployment remains challenging due to strict latency and hardware constraints, as robust temporal reasoning often relies on large backbones and computationally intensive video encoders~\cite{Min2025InnovatingRobotAssistedSurgery}. In this context, knowledge distillation offers a practical way to retain core multimodal capabilities while transferring them into more efficient models for surgical assistance.

\subsection{Distilling Multimodal Large Language Models}
The rapid scaling of MLLMs has made knowledge distillation a primary route toward deployable models without retraining from scratch. Early work on efficient VLMs combines teacher-student distillation with structural sparsification, exemplified by EfficientVLM~\cite{wang-etal-2023-efficientvlm}, which distills a large VLM during pretraining and applies modal-adaptive pruning to remove redundant components. As MLLMs evolve toward instruction-following and open-ended multimodal dialogue, subsequent studies focus on \emph{what} to distill and \emph{where} supervision should be applied. LLAVADI~\cite{xu2024llavadi} systematically analyzes MLLM distillation design choices and emphasizes jointly aligning token-level behavior and output distributions, while CoMD~\cite{li2023unlock} moves beyond one-way imitation by introducing competitive distillation to improve student multimodal competence. To better preserve multimodal reasoning, LLaVA-KD~\cite{cai2025llava} and MoVE-KD~\cite{cao2025movekd} distill cross-modal relations and complementary visual knowledge, respectively, and LLaVA-MoD~\cite{Shu2024LLaVAMoD} further reduces inference cost through sparse MoE design with progressive and preference-aware distillation. Orthogonally, architectural substitutions paired with distillation, such as MaTVLM~\cite{li2025matvlm}, replace parts of the Transformer with Mamba-style layers to recover performance under lower latency.

Beyond objectives and architectures, distillation performance also depends on supervision quality and training dynamics. Self-Improving Teacher improves student transfer by calibrating distillation through challenging samples that refine teacher supervision~\cite{li2024selfimproving}. Importantly, recent work increasingly treats distillation as a \emph{curriculum} rather than a one-shot compression step. Curriculum extraction~\cite{gupta2025curriculumextraction} and progressive consistency distillation~\cite{wen2025epic} demonstrate that staged learning trajectories improve distillation stability and efficiency. Explicit curriculum scheduling has also been explored for knowledge distillation in more generic settings, like CKD-MDSR~\cite{sun2025curriculum}, but remains comparatively underexplored for MLLMs and efficiency-critical domains such as surgical assistance.

\subsection{Advances in Curriculum learning}
Curriculum learning (CL) is a machine learning strategy that arranges training data, tasks, or objectives from simpler to more complex instances to improve convergence and generalization, inspired by human pedagogical principles~\cite{Soviany2022Survey}. Early work formalized CL as a training paradigm where model performance benefits from gradually increasing complexity of examples, and established general frameworks involving difficulty measurers and training schedulers~\cite{Wang2020Curriculum}. Beyond traditional data-level curricula, recent work has extended CL principles to efficient training of large neural architectures. For large language models (LLMs), curriculum distillation frameworks systematically structure training sequences to first solidify basic competencies before introducing harder reasoning tasks, yielding improved performance over single-stage distillation~\cite{jiang2025teach}. In vision-language settings, curriculum strategies have been shown to benefit multimodal evaluation by pacing training samples according to complexity, especially under limited data regimes~\cite{saha2024exploring}. Moreover, curriculum learning has been applied in large-scale language model pretraining, where ordering examples by difficulty metrics consistently improves convergence and early training efficiency compared to random sampling~\cite{zhang2026beyond}. In application-specific domains, specialized curricula designed by domain experts substantially improve clinical disease staging and referral outcomes for medical vision-language models compared to baseline training strategies~\cite{Holland2025SpecCurriculumVLM}. These advances collectively indicate that curriculum learning constitutes a flexible training paradigm that can be adapted for both general efficiency gains and domain-specific performance, motivating its integration with knowledge distillation and adaptive scheduling strategies in surgical MLLMs.

\begin{figure*}[!t]
\centering
\hspace{-0.0cm}
\includegraphics[width=\linewidth]{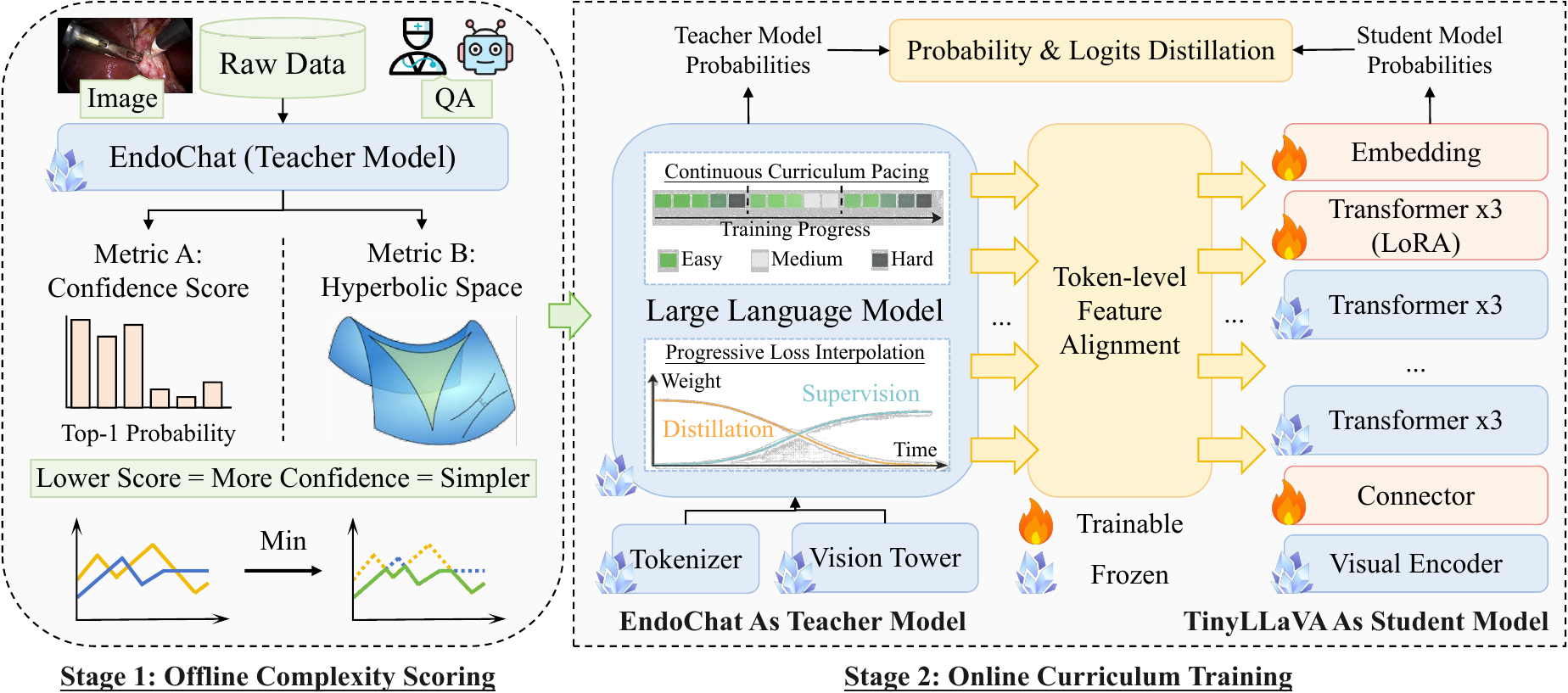}
\caption{\textbf{Overview of the proposed complexity-aware curriculum framework for progressive distillation.}
Teacher confidence and Lorentz geometric consistency are fused into a permissive complexity score, which guides continuous curriculum pacing from easy-first sampling to balanced training. This pacing is coupled with progressive objective interpolation and hierarchical representation alignment, jointly establishing a more structured optimization process that enables stable and efficient compression of surgical vision-language models.}
\label{fig:method}
\end{figure*}

\section{Methods}
\label{sec:methods}
In this section, we present SurgPace, a complexity-aware continuous curriculum framework for \textit{progressive distillation} of surgical vision-language models. The key idea is to estimate the teacher-derived complexity of each training sample, convert this estimate into a continuous easy-to-hard sample pacing mechanism, and couple this sample-level curriculum with hierarchical representation alignment. This formulation explicitly separates two roles that are often entangled in curriculum-based distillation: curriculum determines which samples should receive greater emphasis during training, whereas progressive objective scheduling determines how different supervision signals should be balanced over time. We instantiate SurgPace with EndoChat~\cite{wang2026endochat} as the teacher (13B) and a lightweight multimodal model as the student (3B). The student follows a TinyLLaVA-style architecture built on Phi-2~\cite{javaheripi2023phi}, with a SigLIP~\cite{zhai2023sigmoid} vision encoder and a two-layer GELU-based MLP connector, and is initialized from a pretrained Phi-2-SigLIP checkpoint~\cite{jia2024tinyllava}.
\noindent \textbf{Motivation: Teacher-Informed Complexity Estimation.}
Curriculum learning improves optimization by modulating how training samples are emphasized according to their complexity. In vision-language models, however, complexity is rarely captured by superficial cues such as image appearance or text length alone. Instead, it often arises from semantic ambiguity, cross-modal alignment difficulty, and uncertainty in sequence generation. Existing curriculum strategies therefore commonly rely on heuristic proxies, such as current loss values or metadata-derived priors, which can be noisy, unstable, and highly sensitive to the training dynamics. We adopt a teacher-informed perspective: a sample is considered difficult when a strong teacher model exhibits uncertainty or internal inconsistency while generating the target answer associated with that sample. This view anchors curriculum construction in teacher response behavior for progressive distillation.

\subsection{Overview of SurgPace}
SurgPace integrates two coordinated training mechanisms into a hierarchical distillation base. The distillation base specifies what knowledge is transferred from the teacher to the student, the curriculum mechanism determines which samples should be emphasized at different stages of training, and the progressive objective schedule governs how distillation and supervised objectives are balanced over time. As illustrated in Figure~\ref{fig:method}, each training sample is first assigned a scalar complexity score derived from teacher confidence and Lorentz geometric consistency with respect to the target answer. These precomputed scores are then transformed into progress-dependent sample weights, giving rise to a continuous easy-to-hard pacing strategy whose effect gradually decays as training advances. In parallel, the student is optimized with supervised cross-entropy and hierarchical representation alignment, which jointly enforces distribution alignment, logit matching, representation alignment, and entropy regularization. Their relative contributions are further adjusted through the progressive objective schedule along the training trajectory. In this design, the curriculum acts at the sample level, whereas the objective schedule acts at the loss level. Together, these components make SurgPace a unified and effective framework for stable and efficient progressive distillation of surgical vision-language models.

\subsection{Complexity-Aware Continuous Curriculum}
\label{subsec:curriculum}
Our curriculum is built on two tightly coupled components: a teacher-derived complexity estimator and a continuous pacing function that converts this estimator into optimization weights. A key design choice is to explicitly separate \emph{offline scoring} from \emph{online weighting}. In the surgical MLLM setting, online complexity estimation is highly unstable because batch sizes are small and the student distribution evolves rapidly during training, whereas teacher-derived offline scores remain globally comparable throughout optimization. We therefore compute a scalar score for each training instance once, store it with the dataset, and reuse this fixed signal during training. Under this design, curriculum is not implemented as a heuristic reshuffling of mini-batches, but as a dataset-level prior that induces smooth and progress-dependent sample emphasis.
\\
\\
\noindent \textbf{Complexity Estimator.}
We compute the complexity of each sample offline using a fixed teacher model. The goal is to measure difficulty from the teacher's actual response behavior on the target answer $y_i$, instead of from shallow metadata. For a sample $x_i$ with target sequence length $L$, the teacher produces token probabilities $p^{(T)}_t$ and logits $l^{(T)}_t$ at each decoding step. We then derive two complementary signals. The first asks whether the teacher is locally certain about the next token. The second asks whether the full decoding trajectory remains globally coherent across the answer sequence. These signals capture different failure modes: a teacher may exhibit high confidence at individual steps while still producing an unstable trajectory across the full sequence.
\\
\\
\noindent \textit{Confidence Score ($s_{conf}$):}
This score is defined as the complement of the teacher's mean top-1 probability, and thus serves as a token-level uncertainty estimate in the standard Euclidean probability space:
\begin{equation}
    s_{conf}(x_i) = 1 - \frac{1}{L}\sum_{t=1}^{L} \max_{v} p^{(T)}_t(v).
\end{equation}
Because each term inside the summation lies in $[0,1]$, the resulting score also satisfies $s_{conf}(x_i)\in[0,1]$. A smaller value indicates that the teacher consistently assigns high mass to one dominant token, suggesting that the sample is easy to decode. Conversely, a larger value indicates persistent uncertainty and therefore higher sample complexity. This signal is purely local: it summarizes token-wise certainty in the teacher's standard output space, but does not account for whether the full decoding trajectory remains geometrically coherent across the answer sequence.
\\
\\
\noindent \textit{Lorentz Geometry Score ($s_{\mathrm{lorentz}}$):}
Confidence alone does not describe the stability of the teacher's decoding process over the whole answer. We therefore map teacher logits to the unit Lorentz hyperboloid and measure their dispersion over the answer sequence. For numerical stability, each teacher logit vector is first compressed into a bounded spatial coordinate:
\begin{equation}
    u_t = \frac{l^{(T)}_t}{\sqrt{1 + \|l^{(T)}_t\|^2 + \epsilon}}.
\end{equation}
We then add a time coordinate and lift the vector to the unit Lorentz hyperboloid:
\begin{equation}
    x_t = \left(\sqrt{1+\|u_t\|^2}, u_t\right).
\end{equation}
With the Lorentz inner product defined as
\begin{equation}
    \langle x,y\rangle_L = -x_0y_0 + x_{1:}^{\top}y_{1:},
\end{equation}
each lifted point is valid because
\begin{equation}
    \langle x_t,x_t\rangle_L = -(1+\|u_t\|^2)+\|u_t\|^2=-1.
\end{equation}
The Lorentzian geodesic distance is
\begin{equation}
    d_L(x,y)=\operatorname{arcosh}\!\left(\max\!\left(1+\epsilon,-\langle x,y\rangle_L\right)\right).
\end{equation}
Because a Euclidean average of hyperboloid points need not remain on the manifold, we average only the spatial coordinates and lift the result again:
\begin{equation}
    \bar{u}=\frac{1}{L}\sum_{t=1}^{L}u_t, \qquad
    \bar{x}=\left(\sqrt{1+\|\bar{u}\|^2},\bar{u}\right).
\end{equation}
$\bar{x}$ is obtained by lifting the averaged Euclidean coordinates and is a valid hyperboloid point. The raw geometric dispersion is
\begin{equation}
    s^{\mathrm{raw}}_{\mathrm{lorentz}}(x_i) = \frac{1}{L}\sum_{t=1}^{L} d_L(x_t,\bar{x}).
\end{equation}
The score is small when the teacher follows a compact and stable trajectory, and large when the token trajectory is dispersed or internally inconsistent.

To make the geometric score directly comparable with the confidence score, we normalize it globally over the training set $\mathcal{D}$:
\begin{equation}
    s_{\mathrm{lorentz}}(x_i) =
    \frac{s^{\mathrm{raw}}_{\mathrm{lorentz}}(x_i) - \min_{x \in \mathcal{D}} s^{\mathrm{raw}}_{\mathrm{lorentz}}(x)}
    {\max_{x \in \mathcal{D}} s^{\mathrm{raw}}_{\mathrm{lorentz}}(x) - \min_{x \in \mathcal{D}} s^{\mathrm{raw}}_{\mathrm{lorentz}}(x) + \epsilon}.
\end{equation}
This dataset-level normalization is important in practice. It ensures that the score remains stable and comparable across optimization steps, while effectively preventing the degenerate behavior that could otherwise arise from independently renormalizing scores within each mini-batch.

In the main formulation, we adopt a permissive fusion strategy that combines the two normalized criteria as:
\begin{equation}\label{eq:score_fusion}
    s_i = \min(s_{conf}(x_i), s_{\mathrm{lorentz}}(x_i)).
\end{equation}
This fusion adopts a less restrictive definition of easiness: a sample is assigned a lower complexity score when it appears easy under at least one of the two criteria. In other words, a favorable signal from either confidence or geometric consistency is sufficient to reduce the estimated difficulty.
\\
\\
\noindent \textbf{Continuous Weighting Function.}
After offline scoring, the selected score field is injected into the dataloader and passed to the trainer as sample-level metadata. During optimization, we perform a conversion of the complexity score into an easiness bias term:
\begin{equation}
    b_i = 1 - s_i,
\end{equation}
so that larger values correspond to easier samples. We then linearly map this bias into a bounded interval $[w_{\min}, w_{\max}]$:
\begin{equation}
    w_i = w_{\min} + (w_{\max} - w_{\min}) b_i.
\end{equation}
Under this formulation, easier samples receive larger initial weights, whereas harder samples are relatively down-weighted at the beginning of training. The weights are computed from globally normalized offline scores.
\\
\\
\noindent \textbf{Curriculum Annealing.}
To avoid permanently biasing optimization toward low-complexity samples, we gradually anneal curriculum weights back to unity as training progresses. Let $p \in [0,1]$ denote the normalized training progress. We define an annealing factor:
\begin{equation}
    a(p) = \max\left(0, 1 - \frac{p}{r_{\mathrm{end}}}\right),
\end{equation}
where $r_{\mathrm{end}}$ denotes the point at which explicit curriculum emphasis vanishes. The effective sample weight at progress $p$ can then be formulated as:
\begin{equation}
    w'_i(p) = 1 + a(p)(w_i - 1).
\end{equation}
Thus, $w'_i(p)$ continuously interpolates between the curriculum weight $w_i$ and the unbiased weight $1.0$. When $p$ is small, optimization follows an explicit easy-first regime; when $p \ge r_{\mathrm{end}}$, all samples are treated uniformly.

\subsection{Hierarchical Representation Alignment}
\label{subsec:alignment}
This hierarchical representation alignment objective defines the objective-side core of SurgPace. We formulate teacher-student transfer as a unified multi-level alignment objective under the curriculum-defined sample pacing and the progressive schedule introduced later in Sec.~\ref{subsec:progressive}. Concretely, teacher and student are aligned at three complementary levels: output distributions, pre-softmax response patterns, and latent representations, while all terms share the same token mask, per-sample reduction, and sample-aware weighting scheme. This design is important because multimodal compression errors rarely emerge at only one level. A student may appear correct at the surface output level while already drifting in logit geometry or hidden-state organization, and such hidden inconsistencies are particularly harmful for surgical dialogue and grounding tasks that rely on stable cross-modal reasoning.

\noindent \textit{Trajectory alignment across tokenizers.} The teacher and student use different subword tokenizers, so neither their token indices nor their token counts provide a semantic correspondence. Let $a^{(T)}_t$ and $a^{(S)}_j$ be the positive UTF-8 surface widths of the decoded teacher and student answer tokens, respectively. We locate every token at the center of its normalized span in the decoded text:
\begin{equation}
    r^{(T)}_t=
    \frac{\sum_{q=1}^{t}a^{(T)}_q-\tfrac{1}{2}a^{(T)}_t}
    {\sum_{q=1}^{L_T}a^{(T)}_q},
    \qquad
    r^{(S)}_j=
    \frac{\sum_{q=1}^{j}a^{(S)}_q-\tfrac{1}{2}a^{(S)}_j}
    {\sum_{q=1}^{L_S}a^{(S)}_q}.
\end{equation}
We interpolate the teacher trajectory at the student positions,
\begin{equation}
    \hat l^{(T)}_j=
    \operatorname{Interp}\!\left(
    \{(r^{(T)}_t,l^{(T)}_t)\}_{t=1}^{L_T};r^{(S)}_j
    \right).
\end{equation}
Interpolation is linear between neighboring teacher token centers and uses the nearest endpoint value outside their range. For a single teacher token, its logits are broadcast to all student positions.
For vocabulary alignment, we form a fixed semantic support $\mathcal{A}=\{(v^{(T)}_m,v^{(S)}_m)\}_{m=1}^{M}$ from exact matches between decoded tokens in the two vocabularies. Let $P_T$ and $P_S$ select the corresponding teacher and student coordinates. The aligned logits used by all losses defined on model outputs are
\begin{equation}
    \tilde l^{(T)}_j=e^{\gamma}P_T\hat l^{(T)}_j,
    \qquad
    \hat l^{(S)}_j=P_Sl^{(S)}_j,
\end{equation}
where $\gamma$ is a learned scalar log-scale parameter. This construction restricts comparison to token identities explicitly shared in decoded text.

Let $\mathcal{M}_i$ denote the set of valid next-token positions for sample $i$ after standard causal shifting,  and define the corresponding masked averaging operator over token-level quantities:
\begin{equation}
    \rho_i(\delta) = \frac{1}{|\mathcal{M}_i|}\sum_{t \in \mathcal{M}_i} \delta_{i,t}.
\end{equation}
All distillation losses below are first computed at the token level and then reduced to per-sample losses through $\rho_i(\cdot)$. This reduction matches the actual training behavior and, more importantly, prevents long responses from dominating short but difficult surgical queries, which makes the alignment objective more balanced across samples.
\\
\\
\noindent \textit{Distribution Alignment:} On the shared vocabulary support, we align the student distribution with the projected teacher distribution through temperature-scaled Kullback-Leibler divergence:
\begin{equation}
    \ell^{KL}_i = \rho_i \!\left(
    T^2 \operatorname{KL}\!\left(
    \operatorname{softmax}\!\left(\frac{\tilde{l}^{(T)}_t}{T}\right)
    \,\middle\|\,
    \operatorname{softmax}\!\left(\frac{\hat l^{(S)}_t}{T}\right)
    \right)\right),
\end{equation}
where $T$ is the distillation temperature. This term transfers the teacher's relative token preferences on the shared vocabulary support and provides the student with a fine-grained preference structure, which is especially important for achieving precise, contextually consistent wording in surgical descriptions and question answering.
\\
\\
\noindent \textit{Logit Alignment:} A softmax distribution is invariant to an additive constant at each position. To make the logit term respect the same identifiable response geometry, we define the centering operator $C(z)=z-M^{-1}\mathbf{1}_M\mathbf{1}_M^{\top}z$ and minimize the squared distance between centered logits:
\begin{equation}
    \ell^{L2}_i = \rho_i \!\left(
    \left\| C\!\left(\hat l^{(S)}_t\right)-C\!\left(\tilde{l}^{(T)}_t\right) \right\|_2^2
    \right).
\end{equation}
This removes only the unidentifiable offset while retaining the relative logit pattern and its confidence scale. Whereas $\ell_i^{KL}$ constrains the probability simplex, $\ell_i^{L2}$ matches the centered response geometry before softmax.
\\
\\
\noindent \textit{Latent Alignment:} To complement transfer at the model output with supervision of internal structure, we standardize the centered projected teacher logits, $S(z)=C(z)/\sqrt{M^{-1}\|C(z)\|_2^2+\epsilon}$, before mapping them into the student's hidden dimension through a learnable head $\psi$:
\begin{equation}
    \ell^{feat}_i = \rho_i \!\left(
    \left\|h^{(S)}_{i,t}-\psi\!\left(S\!\left(\tilde{l}^{(T)}_{i,t}\right)\right)\right\|_2^2
    \right),
\end{equation}
The standardization prevents arbitrary differences in teacher logit scale from determining the feature target, while $\psi$ learns the transfer into the student hidden space. This term encourages the student to recover the internal structure provided by the teacher.
\\
\\
\noindent \textit{Entropy Regularization:} Finally, to discourage overly diffuse predictions, we add an entropy-minimization term:
\begin{equation}
    \ell^{ent}_i = \rho_i \!\left( \mathcal{H}(p^{(S)}_{i,t}) \right).
\end{equation}
This term improves prediction sharpness and is particularly useful for open-ended surgical text generation, where under-confident decoding can otherwise quickly accumulate lexical drift over long responses.

Overall, the four losses above jointly instantiate hierarchical representation alignment as a unified multi-level distillation objective. They use the same token mask, per-sample reduction, and curriculum weight before entering the progressive objective schedule. By aligning the student with the teacher at the distribution, logit, and latent levels within the same optimization design, this objective provides a stronger and more reliable foundation for compressed surgical vision-language learning.

\begin{algorithm}[t]
\caption{Complexity-Aware Continuous Curriculum for Progressive Distillation}
\label{alg:training}
\SetAlgoLined
\DontPrintSemicolon
\textbf{Input:} Dataset $\mathcal{D}$, Teacher $T$, Student $S$, Max steps $M$\\
\textbf{Params:} Weights $\lambda = \{w_{ce}, w_{kl}, w_{l2}, w_{feat}, w_{ent}\}$; Hyperparameters $\Theta$\\

\vspace{0.2em}
\For{each sample $(x_i, y_i) \in \mathcal{D}$}{
    Run teacher $T$ on $(x_i, y_i)$ to obtain logits $l^{(T)}$ and probabilities $p^{(T)}$\;
    $s_{conf} \leftarrow \mathcal{S}_{conf}(p^{(T)})$\;
    $s^{\mathrm{raw}}_{\mathrm{lorentz}} \leftarrow \mathcal{S}_{geo}(l^{(T)})$\;
}
$s_{\mathrm{lorentz}} \leftarrow \mathcal{N}(s^{\mathrm{raw}}_{\mathrm{lorentz}})$\;
$s_i \leftarrow \min(s_{conf}, s_{\mathrm{lorentz}})$; store $s_i$ in the dataset\;
$w_i \leftarrow \mathcal{W}(s_i)$\;

\vspace{0.2em}
\For{$step = 1$ \KwTo $M$}{
    Load batch $\mathcal{B}$ with task labels and stored sample scores; set $p \leftarrow step/M$\;
    $w'_i(p) \leftarrow \mathcal{C}(w_i, p)$\;
    $\beta_{ce}(p), \beta_{dist}(p) \leftarrow \mathcal{A}(p)$\;
    $\omega_i(p)\leftarrow\tau_i w'_i(p)$\;

    Student $l^{(S)}, h^{(S)} \leftarrow S(x_i, y_i)$; \quad Teacher $l^{(T)} \leftarrow T(x_i, y_i)$\;
    $\tilde{l}^{(T)} \leftarrow \Pi(l^{(T)})$\;
    $z^{(T)} \leftarrow \psi(\tilde{l}^{(T)})$\;

    $\ell_i^{CE}, \ell_i^{KL}, \ell_i^{L2}, \ell_i^{feat}, \ell_i^{ent} \leftarrow \mathcal{M}(\mathcal{B})$\;
    $\mathcal{L}_{final} \leftarrow \mathcal{R}_p(\ell_i^{CE}, \ell_i^{KL}, \ell_i^{L2}, \ell_i^{feat}, \ell_i^{ent}; \omega_i, \beta_{ce}, \beta_{dist})$\;

    Update student $S$ with $\nabla \mathcal{L}_{final}$\;
}
\end{algorithm}

\subsection{Progressive Distillation Objective}
\label{subsec:progressive}
The final objective unifies the sample-level curriculum above with an objective-level schedule for progressive distillation and supervised learning within a single optimization formulation. These two mechanisms play distinct roles and should be treated separately: curriculum determines \emph{which samples} should matter more at a given stage, whereas the progressive objective schedule determines \emph{which supervision signal} should dominate at that stage. This design makes the interaction between sample pacing and supervision scheduling explicit and more systematically controlled in the final training objective.

Let $p \in [0,1]$ denote the normalized training progress. We define a stage coefficient $\alpha(p)$ by a three-phase schedule with ratios $(r_1,r_2,r_3)$, which governs the transition dynamics across different training stages:
\begin{equation}\label{eq:phase_alpha}
    \alpha(p)=
    \begin{cases}
    0, & p \le r_1,\\[2pt]
    \dfrac{p-r_1}{r_2}, & r_1 < p \le r_1+r_2,\\[8pt]
    1, & p > r_1+r_2.
    \end{cases}
\end{equation}
This coefficient parameterizes the transition from teacher-dominant optimization to task-dominant optimization. Based on this stage coefficient, we further define the corresponding interpolation weights as:
\begin{equation}\label{eq:phase_scales}
    \beta_{dist}(p)=d_s+(d_e-d_s)\alpha(p), \qquad
    \beta_{ce}(p)=c_s+(c_e-c_s)\alpha(p),
\end{equation}
where $(d_s,d_e)$ and $(c_s,c_e)$ denote the start and end scales for distillation and cross-entropy, respectively. Early in training, $\beta_{dist}(p)$ is large and $\beta_{ce}(p)$ is small, so the student first learns to imitate the teacher. Then, the balance shifts toward ground-truth supervision, allowing the student to refine task-specific predictions after a stable multimodal transfer phase.

\begin{figure*}[t]
    \centering
    \includegraphics[width=\textwidth]{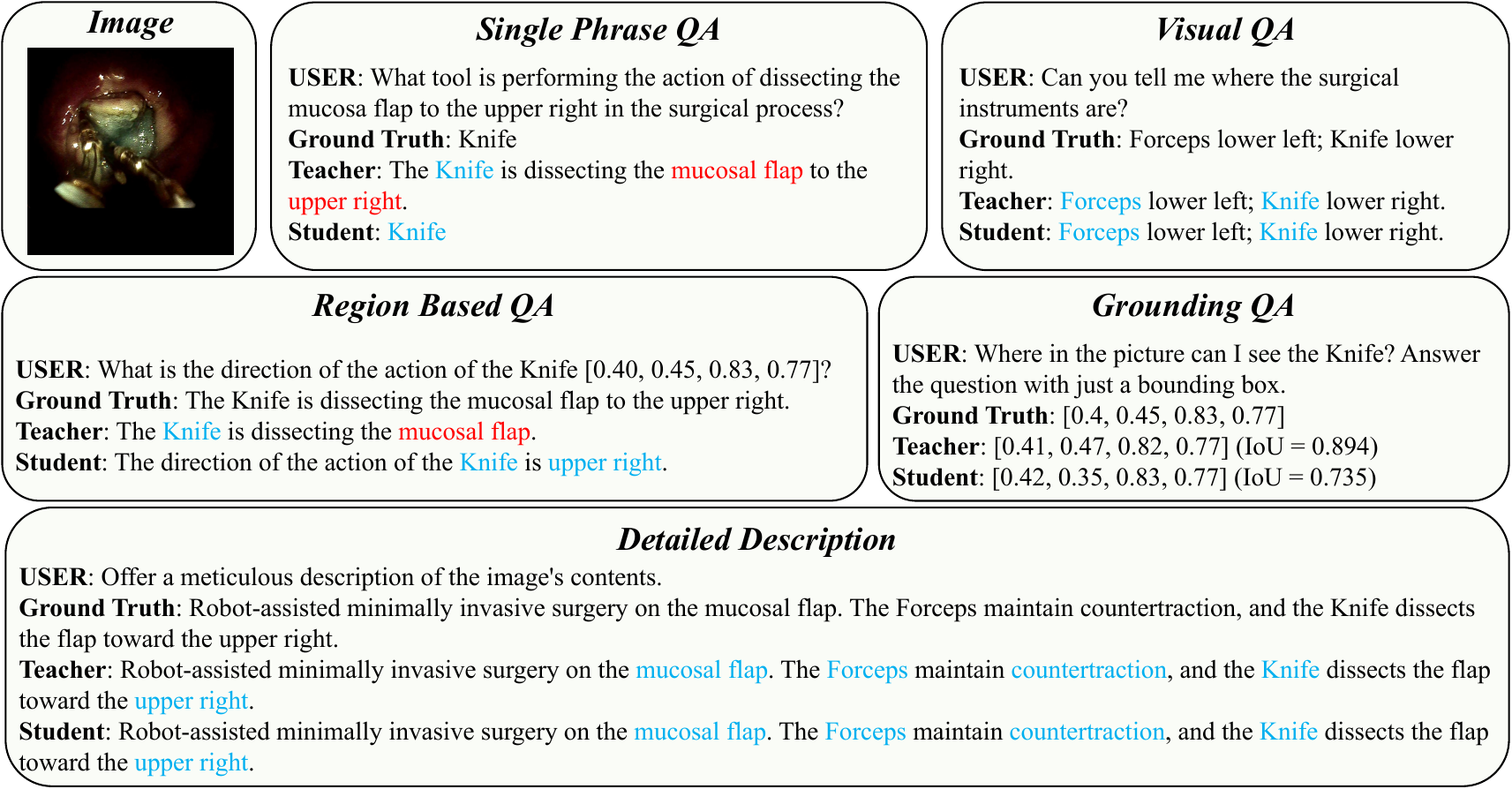}
    \caption{Qualitative visualization of a representative same-image case across five dialogue paradigms. For the same surgical scene, the figure compares ground truth, teacher, and student outputs for single-phrase QA, visual QA, region-based QA, grounding QA, and detailed description. The example shows that the student preserves the main clinical semantics and spatial relationships while remaining compact.}
    \label{fig:visual_case}
\end{figure*}

\begin{figure}[!t]
    \centering
    \includegraphics[width=0.76\linewidth]{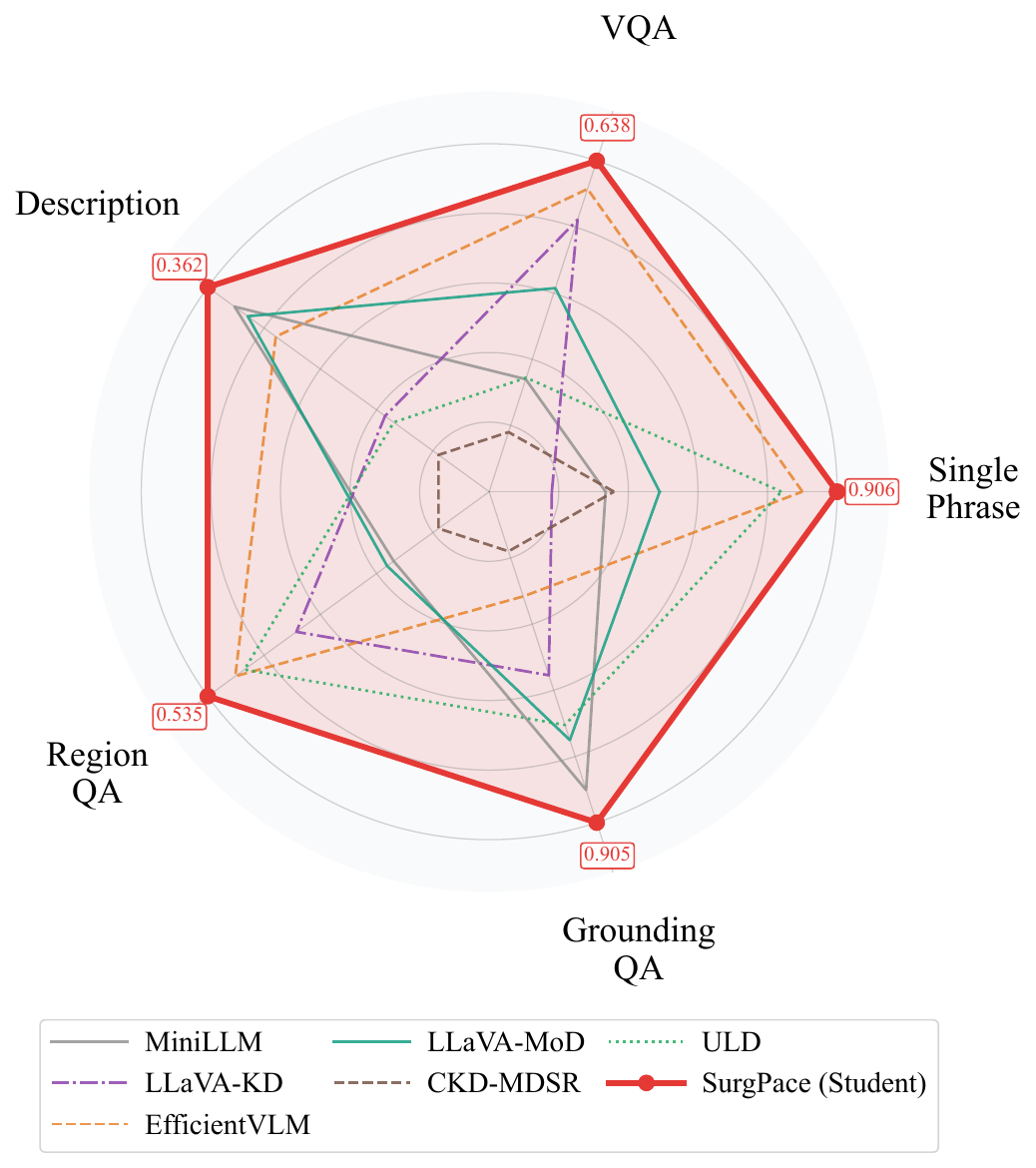}
    \caption{Radar chart of task mean scores computed from Tables~\ref{tab:sp_vqa_results} and~\ref{tab:desc_region_ground_results}. Each axis averages the metrics within one dialogue paradigm.}
    \label{fig:radar_paradigms}
\end{figure}

\begin{figure}[!t]
    \centering
    \captionof{table}{Distribution of the processed Surg-396K split used in this study.}
    \label{tab:data_distribution}
    \setlength{\tabcolsep}{5pt}
    \begin{tabular}{lrr}
    \toprule
    Task family & Training instances & Proportion (\%) \\
    \midrule
    Single-phrase QA   & 156,983 & 34.03 \\
    Visual QA          & 109,960 & 23.84 \\
    Detailed description & 29,960 & 6.49 \\
    Region-based QA    & 92,287  & 20.00 \\
    Grounding QA       & 72,107  & 15.63 \\
    \midrule
    Total              & 461,297 & 100.00 \\
    \bottomrule
    \end{tabular}
\end{figure}

\begin{table*}[t]
\centering
\begin{threeparttable}
\caption{Comparison experiments against state-of-the-art MLLM distillation methods on Single Phrase QA and Visual QA under a fixed 64,856-example evaluation source and greedy decoding.}
\label{tab:sp_vqa_results}
\small
\setlength{\tabcolsep}{4pt}
\renewcommand{\arraystretch}{1.08}
\begin{tabular}{@{}l*{9}{c}@{}}
\toprule
\multirow{2}{*}{\textbf{Method}}
& \multicolumn{4}{c}{\textbf{Single Phrase QA}}
& \multicolumn{5}{c}{\textbf{Visual QA}} \\
\cmidrule(lr){2-5} \cmidrule(lr){6-10}
& Acc.  & Prec.  & Rec.  & F1
& BLEU-3  & CIDEr  & METEOR  & ROUGE-1  & ROUGE-L  \\
\midrule
MiniLLM~\cite{gu2024minillm} & 0.8265 & 0.8384 & 0.8446 & 0.8402 & 0.3246 & 0.4676 & 0.7608 & 0.7521 & 0.7083 \\
EfficientVLM~\cite{wang-etal-2023-efficientvlm} & 0.8296 & \underline{0.8412} & \underline{0.8475} & \underline{0.8431} & \textbf{0.3801} & \underline{0.4830} & 0.7705 & \underline{0.7632} & \underline{0.7204} \\
ULD~\cite{boizard2024towards} & \underline{0.8307} & 0.8406 & 0.8458 & 0.8421 & 0.3215 & 0.4683 & 0.7613 & 0.7539 & 0.7098 \\
LLaVA-KD~\cite{cai2025llava} & 0.8241 & 0.8361 & 0.8421 & 0.8378 & 0.3689 & 0.4797 & 0.7686 & 0.7599 & 0.7152 \\
LLaVA-MoD~\cite{Shu2024LLaVAMoD} & 0.8279 & 0.8388 & 0.8466 & 0.8415 & 0.3582 & 0.4649 & \underline{0.7757} & 0.7599 & 0.7000 \\
CKD-MDSR~\cite{sun2025curriculum} & 0.8269 & 0.8385 & 0.8448 & 0.8404 & 0.0520 & 0.3074 & 0.5800 & 0.6026 & 0.5492 \\
\rowcolor{yellow!15}
\textbf{Ours (Student)} & \textbf{0.9004} & \textbf{0.9067} & \textbf{0.9092} & \textbf{0.9069} & \underline{0.3764} & \textbf{0.5135} & \textbf{0.7841} & \textbf{0.7809} & \textbf{0.7366} \\\hline
EndoChat (Teacher)\textsuperscript{a}~\cite{wang2026endochat} & 0.5485 & 0.6046 & 0.7647 & 0.6325 & 0.3941 & 0.5067 & 0.7967 & 0.7869 & 0.7414 \\
\bottomrule
\end{tabular}
\begin{tablenotes}[flushleft]
\footnotesize
\item[a] EndoChat is reported as the teacher reference and is excluded from best/second-best marking.
\item The best and second-best results among the student methods are \textbf{bolded} and \underline{underlined}, respectively.
\item All student rows use the common conversation template and generation protocol described in Section~4.2, including greedy decoding with \texttt{num\_beams=1}, \texttt{max\_new\_tokens=256}, and the same output normalization. No prompt augmentation for individual regions or task-specific decoding constraint is used; EndoChat is reported as the teacher reference.
\end{tablenotes}
\end{threeparttable}
\end{table*}

\begin{table*}[t]
\centering
\begin{threeparttable}
\caption{Comparison experiments against state-of-the-art MLLM distillation methods on Detailed Description, Region-Based QA, and Grounding QA under the same fixed evaluation protocol as Table~\ref{tab:sp_vqa_results}.}
\label{tab:desc_region_ground_results}
\small
\setlength{\tabcolsep}{4pt}
\renewcommand{\arraystretch}{1.08}
\begin{tabular}{@{}l*{7}{c}@{}}
\toprule
\multirow{2}{*}{\textbf{Method}}
& \textbf{Detailed Description}
& \multicolumn{4}{c}{\textbf{Region-Based QA}}
& \multicolumn{2}{c}{\textbf{Grounding QA}} \\
\cmidrule(lr){2-2} \cmidrule(lr){3-6} \cmidrule(lr){7-8}
& ROUGE-L
& CIDEr  & METEOR  & ROUGE-1  & ROUGE-L
& AP50  & mIoU  \\
\midrule
MiniLLM~\cite{gu2024minillm} & \underline{0.3345} & 0.2931 & 0.5782 & 0.5964 & 0.5566 & \underline{0.5857} & \underline{0.5104} \\
EfficientVLM~\cite{wang-etal-2023-efficientvlm} & 0.3284 & 0.3034 & 0.5875 & \underline{0.6038} & \underline{0.5661} & 0.5454 & 0.4887 \\
ULD~\cite{boizard2024towards} & 0.3134 & 0.3038 & \underline{0.5921} & 0.6013 & 0.5600 & 0.5708 & 0.5022 \\
LLaVA-KD~\cite{cai2025llava} & 0.3154 & 0.3046 & 0.5918 & 0.5978 & 0.5520 & 0.5636 & 0.4993 \\
LLaVA-MoD~\cite{Shu2024LLaVAMoD} & 0.3317 & \underline{0.3053} & 0.5786 & 0.5943 & 0.5493 & 0.5742 & 0.5019 \\
CKD-MDSR~\cite{sun2025curriculum} & 0.1157 & 0.2277 & 0.4845 & 0.5183 & 0.4844 & 0.4946 & 0.4563 \\
\rowcolor{yellow!15}
\textbf{Ours (Student)} & \textbf{0.3621} & \textbf{0.3311} & \textbf{0.6010} & \textbf{0.6217} & \textbf{0.5878} & \textbf{0.9625} & \textbf{0.8471} \\ \hline
EndoChat (Teacher)\textsuperscript{a}~\cite{wang2026endochat} & 0.3307 & 0.3930 & 0.7049 & 0.7119 & 0.6659 & 0.9664 & 0.8789 \\
\bottomrule
\end{tabular}
\begin{tablenotes}[flushleft]
\footnotesize
\item[a] EndoChat is reported as the teacher reference and is excluded from best/second-best marking.
\item The best and second-best results among the student methods are \textbf{bolded} and \underline{underlined}, respectively.
\item The values are evaluated under the fixed protocol stated in Table~\ref{tab:sp_vqa_results}.
\end{tablenotes}
\end{threeparttable}
\end{table*}

\begin{table*}[t]
\centering
\begin{threeparttable}
\caption{Robustness comparison across three seeds between SurgPace and the strongest baseline, EfficientVLM, using the same ten task-spanning metrics reported in the ablation studies. Each cell gives the mean, with the sample standard deviation shown below, over seeds 0, 1, and 42 under the fixed common evaluation protocol.}
\label{tab:multiseed_results}
\small
\setlength{\tabcolsep}{3pt}
\renewcommand{\arraystretch}{1.08}
\begin{tabular}{@{}l*{10}{c}@{}}
\toprule
\multirow{2}{*}{\textbf{Method}}
& \multicolumn{2}{c}{\textbf{Single Phrase QA}}
& \multicolumn{3}{c}{\textbf{Visual QA}}
& \textbf{Description}
& \multicolumn{2}{c}{\textbf{Region-Based QA}}
& \multicolumn{2}{c}{\textbf{Grounding QA}} \\
\cmidrule(lr){2-3} \cmidrule(lr){4-6} \cmidrule(lr){7-7} \cmidrule(lr){8-9} \cmidrule(lr){10-11}
& Acc. & F1 & BLEU-3 & CIDEr & METEOR & ROUGE-L & CIDEr & METEOR & AP50 & mIoU \\
\midrule
EfficientVLM
& \makecell{$0.8312$\\{\scriptsize $\pm0.0014$}} & \makecell{$0.8447$\\{\scriptsize $\pm0.0014$}}
& \makecell{$0.3741$\\{\scriptsize $\pm0.0056$}} & \makecell{$0.4850$\\{\scriptsize $\pm0.0018$}} & \makecell{$0.7705$\\{\scriptsize $\pm0.0016$}}
& \makecell{$0.3259$\\{\scriptsize $\pm0.0057$}}
& \makecell{$0.3130$\\{\scriptsize $\pm0.0084$}} & \makecell{$0.5871$\\{\scriptsize $\pm0.0007$}}
& \makecell{$0.5686$\\{\scriptsize $\pm0.0229$}} & \makecell{$0.5032$\\{\scriptsize $\pm0.0136$}} \\
\rowcolor{yellow!15}
\textbf{Ours (Student)}
& \makecell{$\mathbf{0.8915}$\\{\scriptsize $\mathbf{\pm0.0101}$}} & \makecell{$\mathbf{0.8990}$\\{\scriptsize $\mathbf{\pm0.0092}$}}
& \makecell{$\mathbf{0.3822}$\\{\scriptsize $\mathbf{\pm0.0050}$}} & \makecell{$\mathbf{0.5119}$\\{\scriptsize $\mathbf{\pm0.0014}$}} & \makecell{$\mathbf{0.7843}$\\{\scriptsize $\mathbf{\pm0.0042}$}}
& \makecell{$\mathbf{0.3589}$\\{\scriptsize $\mathbf{\pm0.0033}$}}
& \makecell{$\mathbf{0.3292}$\\{\scriptsize $\mathbf{\pm0.0038}$}} & \makecell{$\mathbf{0.5999}$\\{\scriptsize $\mathbf{\pm0.0016}$}}
& \makecell{$\mathbf{0.9604}$\\{\scriptsize $\mathbf{\pm0.0031}$}} & \makecell{$\mathbf{0.8424}$\\{\scriptsize $\mathbf{\pm0.0069}$}} \\
\bottomrule
\end{tabular}
\begin{tablenotes}[flushleft]
\footnotesize
\item The higher mean between the two student methods is \textbf{bolded}; the final model is highlighted in yellow.
\item The 16-metric mean, computed independently for each seed and then summarized, is $0.6221\pm0.0029$ for EfficientVLM and $0.6930\pm0.0038$ for SurgPace.
\end{tablenotes}
\end{threeparttable}
\end{table*}

\begin{table}[!t]
\centering
\begin{threeparttable}
\caption{Sensitivity to perturbations of the teacher logits used only for offline difficulty scoring. All settings use EndoChat, seed 42, identical online distillation, and the same training and common evaluation protocols.}
\label{tab:teacher_sensitivity}
\footnotesize
\setlength{\tabcolsep}{3pt}
\renewcommand{\arraystretch}{1.08}
\begin{tabular}{@{}lcc@{}}
\toprule
\textbf{Offline score setting} & \textbf{16-metric avg.} & \textbf{Change} \\
\midrule
\rowcolor{yellow!15}
Original teacher logits & 0.6955 & -- \\
Temperature calibration ($T=1.3$) & 0.6926 & $-0.0029$ ($-0.4\%$) \\
Noise on centered logits ($\sigma_{\mathrm{rel}}=0.05$) & 0.6887 & $-0.0068$ ($-1.0\%$) \\
\bottomrule
\end{tabular}
\begin{tablenotes}[flushleft]
\footnotesize
\item The configuration using the original logits is the final setting and is highlighted in yellow.
\end{tablenotes}
\end{threeparttable}
\vspace{6pt}

\centering
\begin{threeparttable}
\caption{Uniform LLM-as-a-Judge evaluation of all Detailed Description outputs using Qwen-3.8-Max at temperature 0. Each system is evaluated on all 3,747 Detailed Description samples.}
\label{tab:llm_judge}
\small
\setlength{\tabcolsep}{3pt}
\renewcommand{\arraystretch}{1.08}
\begin{tabular}{@{}l*{3}{c}@{}}
\toprule
\textbf{Method} & \textbf{LLM Score $\uparrow$} & \textbf{Std} & \textbf{95\% CI} \\
\midrule
MiniLLM & 60.73 & 18.81 & [60.12, 61.33] \\
EfficientVLM & \underline{68.98} & 20.35 & [68.33, 69.60] \\
ULD & 58.52 & 26.92 & [57.71, 59.42] \\
LLaVA-KD & 59.36 & 26.93 & [58.51, 60.22] \\
LLaVA-MoD & 67.47 & 18.89 & [66.88, 68.07] \\
CKD-MDSR & 1.77 & 6.21 & [1.58, 1.97] \\
\rowcolor{yellow!15}
\textbf{Ours (Student)} & \textbf{77.48} & 15.02 & [77.00, 77.96] \\ \hline
EndoChat (Teacher)\textsuperscript{a} & 76.67 & 13.38 & [76.24, 77.09] \\
\bottomrule
\end{tabular}
\begin{tablenotes}[flushleft]
\footnotesize
\item[a] EndoChat is reported as the teacher reference and is excluded from best/second-best marking.
\item The best and second-best LLM Scores among the student methods are \textbf{bolded} and \underline{underlined}, respectively.
\item The judge receives only the reference and generated text; it does not receive the image or model identity. All methods use the same EndoChat prompt, judge, temperature, seed, and sample set. All scores in this column are generated with this single configuration.
\end{tablenotes}
\end{threeparttable}
\end{table}

\begin{table}[!t]
\centering
\begin{threeparttable}
\caption{Blinded clinical audit based on reference text across six dimensions using Qwen-3.8-Max on 300 shared stratified Detailed Description samples per system. Scores range from 0 to 4; higher is better, and a hallucination score of 4 denotes no unsupported content.}
\label{tab:clinical_audit}
\footnotesize
\setlength{\tabcolsep}{2.5pt}
\renewcommand{\arraystretch}{1.08}
\begin{tabular}{@{}p{0.31\columnwidth}
>{\centering\arraybackslash}p{0.19\columnwidth}
>{\columncolor{yellow!15}\centering\arraybackslash}p{0.19\columnwidth}
>{\centering\arraybackslash}p{0.19\columnwidth}@{}}
\toprule
\textbf{Audit dimension} & \makecell{\textbf{EfficientVLM}} & \makecell{\textbf{Ours}\\\textbf{(Student)}} & \makecell{\textbf{EndoChat}\\\textbf{(Teacher)}\textsuperscript{a}} \\
\midrule
Overall (0--100) & \underline{64.92} & \textbf{72.70} & 63.97 \\
Instrument & \underline{3.02} & \textbf{3.40} & 3.66 \\
Tissue & \underline{2.91} & \textbf{3.19} & 2.79 \\
Action/motion & \textbf{2.65} & \underline{2.62} & 1.36 \\
Direction & \underline{2.24} & \textbf{2.45} & 1.75 \\
Spatial & \underline{2.29} & \textbf{2.91} & 2.61 \\
Hallucination & \underline{2.52} & \textbf{2.93} & 3.23 \\
\bottomrule
\end{tabular}
\begin{tablenotes}[flushleft]
\footnotesize
\item[a] EndoChat is reported as the teacher reference and is excluded from best/second-best marking.
\item The best and second-best audit scores among the student methods are \textbf{bolded} and \underline{underlined}, respectively.
\end{tablenotes}
\end{threeparttable}
\end{table}

\begin{table}[!t]
\centering
\begin{threeparttable}
\caption{Flagged error counts from the shared blinded clinical audit. Counts are reported out of the same 300 stratified Detailed Description samples for each system.}
\label{tab:clinical_audit_errors}
\footnotesize
\setlength{\tabcolsep}{2.5pt}
\renewcommand{\arraystretch}{1.08}
\begin{tabular}{@{}p{0.34\columnwidth}
>{\centering\arraybackslash}p{0.18\columnwidth}
>{\columncolor{yellow!15}\centering\arraybackslash}p{0.18\columnwidth}
>{\centering\arraybackslash}p{0.18\columnwidth}@{}}
\toprule
\textbf{Flagged error type} & \makecell{\textbf{Efficient}\\\textbf{VLM}} & \makecell{\textbf{Ours}\\\textbf{(Student)}} & \makecell{\textbf{EndoChat}\\\textbf{(Teacher)}} \\
\midrule
\makecell[l]{Instrument identity/\\count} & 127 & 83 & 46 \\
\makecell[l]{Tissue/target\\relationship} & 87 & 70 & 109 \\
Action/motion & 110 & 151 & 273 \\
Direction & 218 & 183 & 227 \\
\makecell[l]{Spatial grounding/\\relation} & 233 & 132 & 143 \\
\makecell[l]{Hallucination/\\unsupported content} & 171 & 151 & 92 \\
\bottomrule
\end{tabular}
\begin{tablenotes}[flushleft]
\footnotesize
\item The same 300 source indices are used for all systems and are sampled by dataset and quartile of reference length within each dataset. Future clinical validation will use adjudication by clinicians with access to the original image.
\end{tablenotes}
\end{threeparttable}
\vspace{6pt}

\centering
\begin{threeparttable}
\caption{Representative SurgPace cases with low scores selected from the shared clinical audit using the predefined error dimensions.}
\label{tab:failure_cases}
\footnotesize
\setlength{\tabcolsep}{2pt}
\renewcommand{\arraystretch}{1.08}
\begin{tabular}{@{}>{\raggedright\arraybackslash}p{0.96\columnwidth}@{}}
\toprule
\textbf{Case and reference--generation discrepancy} \\
\midrule
\textbf{CoPESD\_000937\_0697}\newline
\textit{Flagged dimensions: Instrument; hallucination.}\newline
The reference identifies forceps and a knife as the two primary objects. The generation states that three main objects are present and adds unsupported scene framing; the main actions are preserved but spatial precision is reduced. \\
\addlinespace
\textbf{CoPESD\_001815\_0240}\newline
\textit{Flagged dimensions: Tissue; action; direction; spatial.}\newline
The reference places the forceps centrally and the knife at the lower right, dissecting toward the lower left. The generation moves the forceps to the left and changes the knife motion to upward and to the right, altering both the action direction and spatial relation. \\
\addlinespace
\textbf{CoPESD\_230943\_0082}\newline
\textit{Flagged dimension: Hallucination.}\newline
The reference explicitly states that the image is too unclear to confirm any object. The generation nevertheless specifies a mucosal flap, a knife, coordinates, and an upward dissection action. \\
\bottomrule
\end{tabular}
\end{threeparttable}
\end{table}

\begin{table}[!t]
\centering
\begin{threeparttable}
\caption{Controlled text generation efficiency on one NVIDIA A100 80GB GPU. The 13B teacher and 3B student are both evaluated in FP16 for a precision-matched comparison; both use batch size 1, 128 input tokens, 64 output tokens, greedy decoding, three warm-up runs, and ten measured repeats.}
\label{tab:efficiency}
\footnotesize
\setlength{\tabcolsep}{2pt}
\renewcommand{\arraystretch}{1.08}
\begin{tabular}{@{}>{\raggedright\arraybackslash}p{0.34\columnwidth}
>{\centering\arraybackslash}p{0.18\columnwidth}
>{\centering\arraybackslash}p{0.18\columnwidth}
>{\centering\arraybackslash}p{0.20\columnwidth}@{}}
\toprule
\textbf{Metric} & \makecell{\textbf{13B Teacher}\\\textbf{(FP16)}} & \makecell{\textbf{3B Student}\\\textbf{(FP16)}} & \makecell{\textbf{Reduction}\\\textbf{/ speedup}} \\
\midrule
Parameters & 13.02B & 3.22B & 4.04$\times$ fewer \\
Active text parameters & 13.02B & 2.78B & 4.69$\times$ fewer \\
Request FLOP proxy for text backbone & $5.00\!\times\!10^{12}$ & $1.07\!\times\!10^{12}$ & 4.68$\times$ fewer \\
Loaded GPU memory & 26.03 GB & 6.46 GB & 4.03$\times$ less \\
Peak GPU memory & 26.20 GB & 6.52 GB & 4.02$\times$ less \\
Prefill / TTFT & 31.95 ms & 18.55 ms & 1.72$\times$ faster \\
Median decode latency & 23.51 ms/token & 16.50 ms/token & 1.42$\times$ faster \\
P95 decode latency & 25.28 ms/token & 21.55 ms/token & 1.17$\times$ faster \\
Output throughput & 42.22 token/s & 58.53 token/s & 1.39$\times$ higher \\
End-to-end latency & 1,547.83 ms & 1,111.67 ms & 1.39$\times$ faster \\
Input throughput & 4,005.97 token/s & 6,899.23 token/s & 1.72$\times$ higher \\
Total throughput & 124.04 token/s & 172.71 token/s & 1.39$\times$ higher \\
\bottomrule
\end{tabular}
\begin{tablenotes}[flushleft]
\footnotesize
\item The FLOP proxy uses $2P$ per token for active text parameters and excludes quadratic attention, image encoding, and sampling overhead.
\end{tablenotes}
\end{threeparttable}
\end{table}

For sample $i$, we further define a combined coefficient:
\begin{equation}
    \omega_i(p) = \tau_i \, w'_i(p),
\end{equation}
where $\tau_i$ is an optional task-level reweighting factor and $w'_i(p)$ is the curriculum weight from Eq.~(14). Let $\mathcal{B}_{\mathrm{global}}$ denote the global mini-batch and $N_{\mathrm{global}}=|\mathcal{B}_{\mathrm{global}}|$ its number of samples. We use the following weighted reduction:
\begin{equation}
    \mathcal{R}^{\mathrm{global}}_p
    = \frac{1}{N_{\mathrm{global}}}
    \sum_{i\in\mathcal{B}_{\mathrm{global}}}
    \omega_i(p)\ell_i.
\end{equation}
Using the unweighted global sample count prevents curriculum weights from being normalized away by mini-batch composition. These token-level losses are first reduced to per-sample quantities and then aggregated with sample-aware weights. The supervised term is given by the token-level cross-entropy:
\begin{equation}
    \ell^{CE}_i = \rho_i\!\left(-\log p^{(S)}_{i,t}(y_{i,t})\right),
\end{equation}
and the final objective becomes:
\begin{equation}
\label{eq:final_objective}
\begin{split}
    \mathcal{L}_{final}(p)
    = \beta_{ce}(p)\, w_{ce}\,\mathcal{R}^{\mathrm{global}}_p(\{\ell^{CE}_i\})
    + \beta_{dist}(p)\Big(
    w_{kl}\mathcal{R}^{\mathrm{global}}_p(\{\ell^{KL}_i\}) \\
    + w_{l2}\mathcal{R}^{\mathrm{global}}_p(\{\ell^{L2}_i\})
    + w_{feat}\mathcal{R}^{\mathrm{global}}_p(\{\ell^{feat}_i\})
    + w_{ent}\mathcal{R}^{\mathrm{global}}_p(\{\ell^{ent}_i\}) \Big),
\end{split}
\end{equation}
where $w_{(\cdot)}$ are fixed hyperparameters. In all experiments, we set $(w_{ce}, w_{kl}, w_{l2}, w_{feat}, w_{ent}) = (0.5, 1.0, 0.5, 0.2, 0.05)$, and analyze the contribution of the corresponding alignment components in Table~\ref{tab:ablation_loss}. Under this formulation, curriculum changes the relative contribution of samples, whereas the progressive objective schedule changes the relative contribution of objectives. One controls which samples dominate optimization, and the other controls which supervision signals dominate at each stage of training. This explicit separation of roles yields a stable framework for progressive distillation of surgical MLLMs. The complete training procedure of SurgPace is summarized in Algorithm~\ref{alg:training}.

\section{Experiments}
\label{sec:results}
\subsection{Dataset}
All experiments were conducted on the Surg-396K surgical multimodal instruction benchmark~\cite{wang2026endochat}. The benchmark is derived from four public datasets, namely CoPESD~\cite{wang2025copesd}, Cholec80-VQA~\cite{seenivasan2022surgicalvqa}, EndoVis17-VQLA, and EndoVis18-VQLA~\cite{bai2023catvil}.
The open-source benchmark provides various tasks in surgical scenes, including spatial grounding, instrument recognition, instrument counting, object position, target tissue, fine-grained motion, and direction, enabling the benchmark to cover both semantic understanding and localized reasoning. We evaluate SurgPace on five dialogue paradigms (single-phrase QA, visual QA, detailed description, region-based QA, and grounding QA, as presented in Table~\ref{tab:data_distribution}) to assess whether a compact surgical MLLM can retain broad multimodal capability after distillation.
For all experiments, we use a fixed processed split consisting of 461,297 training instances, 64,878 validation instances, and 64,856 test instances, covering 591,031 instruction-tuning QA pairs in total.
The same split is used for all baselines, ablations, and final comparisons.

\subsection{Implementation Details}
All student methods, including SurgPace and the comparison methods, use the same 3B TinyLLaVA-style student architecture built on Phi-2 with a SigLIP vision tower and a two-layer GELU-based multimodal connector. All methods use the same pretrained Phi-2--SigLIP student initialization and the same EndoChat teacher for distillation.
For parameter-efficient tuning, all student methods use LoRA on the language model, keep the vision tower frozen, and fully fine-tune the multimodal connector. The shared LoRA configuration uses rank 64, scaling factor $\alpha=128$, and \texttt{bias=None}. LoRA targets \texttt{q\_proj}, \texttt{k\_proj}, \texttt{v\_proj}, \texttt{dense}, \texttt{mlp.fc1}, and \texttt{mlp.fc2} in the 32 language-model layers, giving 192 target modules. Thus, the parameter-efficient fine-tuning setup is shared; the method-specific differences arise mainly from the distillation objectives and training strategies.
A lightweight learnable projection layer is introduced for teacher-student logit alignment. Input images are processed under a square aspect-ratio policy, visual features are extracted from the penultimate vision layer, and the maximum sequence length is set to 1024 tokens.
All student methods use BF16 with DeepSpeed ZeRO-2~\cite{rajbhandari2020zero} and the same AdamW optimizer~\cite{loshchilov2017decoupled}, learning rate $1\times10^{-4}$, cosine decay, and warm-up ratio 0.03. The main experiments are run for one epoch with a per-device batch size of 1 on two NVIDIA A100 80GB GPUs.
The fixed processed split contains 461,297 training instances, 64,878 validation instances, and 64,856 test instances, and is used for all comparison methods, ablations, and final evaluations. Fine-tuning strategies differ by method, but all methods are based on supervised fine-tuning (SFT). Training budgets are not identical. Table~\ref{tab:training_budget} reports the per-method GPU-hours; for SurgPace, the reported total includes 52.443 GPU-hours of training and 3.11 GPU-hours of offline difficulty scoring. Across all seven compared methods, the mean total is 71.572 GPU-hours, with a maximum of 165.514 GPU-hours for LLaVA-KD and a minimum of 37.890 GPU-hours for ULD.
For comparison, we evaluate against MiniLLM~\cite{gu2024minillm}, EfficientVLM~\cite{wang-etal-2023-efficientvlm}, ULD~\cite{boizard2024towards}, LLaVA-KD~\cite{cai2025llava}, LLaVA-MoD~\cite{Shu2024LLaVAMoD}, and CKD-MDSR~\cite{sun2025curriculum}, and report EndoChat~\cite{wang2026endochat} as the teacher reference.
For the common generation protocol, all student methods use the same original questions and the common conversation template used in the TinyLLaVA-based implementation~\cite{jia2024tinyllava}. The shared settings include automatic insertion of one \texttt{<image>} token, greedy decoding with \texttt{num\_beams=1} and \texttt{max\_new\_tokens=256}, \texttt{skip\_special\_tokens=True}, and identical whitespace and terminal stop-string handling.
At the metric stage, outputs are lowercased, \texttt{<image>} is replaced with a space, and the same token normalization is applied before exact-match and overlap evaluation; ROUGE, BLEU/CIDEr/METEOR, and Grounding use the same respective evaluation procedures for all methods.
The complexity scores are computed once offline using forward passes through the teacher. Scoring all 591,031 instances required 3.11 GPU hours, and the resulting scores are reused throughout training.

For semantic evaluation of Detailed Description, we use the original EndoChat LLM-as-a-Judge prompt\footnote{``You have been given two detailed descriptions of a surgical scene, the first one is the ground truth (enclosed within the \texttt{<ground\_truth></ground\_truth>} tags) and the second one is
the detailed description generated by a large language model (enclosed within
the \texttt{<generation></generation>} tags). I want you to be a surgeon who
specializes in operating surgical robots. Please score the accuracy
of the detailed description generated by the large language model
based on the ground truth in terms of accuracy and quality. The
score should be an integer ranging from 0 to 100. Please only
output the score.''} and evaluate all eight systems, including
EndoChat itself, with Qwen-3.8-Max. For each of the 3,747 Detailed Description test samples, the judge receives only the reference and generated text, is asked to assess accuracy and quality as a specialist in surgical robotics, and returns one integer from 0 to 100. We use temperature 0 and seed 42, and report the mean, sample standard deviation, and a 2,000-resample bootstrap 95\% confidence interval. We separately conduct a blinded audit based on reference text across six dimensions using 300 shared samples stratified by dataset and quartile of reference length. The audit scores instrument identity/count, tissue/target relationship, action/motion, direction, spatial grounding/relations, and hallucination from 0 to 4; model identity is hidden and verbosity is not rewarded.
The complete blinded-audit prompt, JSON output schema, scoring anchors, and recorded run configuration are provided in Supplementary Material, Section~S1.

We additionally benchmark controlled text generation on one NVIDIA A100 80GB GPU. The 13B teacher and 3B student are both evaluated in FP16 for a precision-matched comparison. Both use batch size 1, 128 input tokens, 64 generated tokens, and greedy decoding. After three warm-up runs, we report the median over ten measured runs. The analytical FLOP proxy is $2P$ per token for the active text parameters and excludes the attention quadratic term, image encoder, and sampling overhead.

Unless otherwise stated, every ablation row is trained with seed 42 under the same implementation, data split, and common evaluation protocol; each row changes only the factor named in its table.

\begin{table*}[t]
\centering
\begin{threeparttable}
\caption{Ablation on \textbf{complexity score fusion} in the offline scoring stage. The variants correspond to Eq.~\eqref{eq:score_fusion}, which fuses the teacher-derived confidence score $s_{\mathrm{conf}}$ computed in the standard Euclidean probability space and the Lorentz geometric score $s_{\mathrm{lorentz}}$ into the final complexity score $s_i$. The adopted final configuration is highlighted in yellow.}
\label{tab:ablation_fusion}
\small
\setlength{\tabcolsep}{3pt}
\renewcommand{\arraystretch}{1.08}
\begin{tabular}{@{}l*{10}{c}@{}}
\toprule
\multirow{2}{*}{\textbf{Variant}}
& \multicolumn{2}{c}{\textbf{Single Phrase QA}}
& \multicolumn{3}{c}{\textbf{Visual QA}}
& \textbf{Detailed Description}
& \multicolumn{2}{c}{\textbf{Region-Based QA}}
& \multicolumn{2}{c}{\textbf{Grounding QA}} \\
\cmidrule(lr){2-3} \cmidrule(lr){4-6} \cmidrule(lr){7-7} \cmidrule(lr){8-9} \cmidrule(lr){10-11}
& Acc.  & F1  & BLEU-3  & CIDEr  & METEOR  & ROUGE-L  & CIDEr  & METEOR  & AP50  & mIoU  \\
\midrule
$s_{\mathrm{conf}}$          & 0.8826 & 0.8908 & 0.3662 & 0.5102 & 0.7799 & 0.3548 & 0.3260 & 0.5984 & 0.9595 & 0.8350 \\
$s_{\mathrm{lorentz}}$       & \underline{0.8997} & \underline{0.9066} & 0.3737 & 0.5099 & 0.7817 & 0.3587 & 0.3257 & 0.5988 & 0.9614 & 0.8420 \\
$s_{\mathrm{max}}$           & 0.8925 & 0.8992 & 0.3741 & 0.5109 & 0.7821 & 0.3561 & \underline{0.3296} & \textbf{0.6018} & 0.9618 & 0.8390 \\
$s_{\mathrm{mean}}$          & 0.8971 & 0.9041 & \underline{0.3754} & \underline{0.5111} & \underline{0.7834} & \underline{0.3591} & 0.3289 & \underline{0.6017} & \underline{0.9624} & \underline{0.8439} \\
\rowcolor{yellow!15}
$s_{\mathrm{min}}$           & \textbf{0.9004} & \textbf{0.9069} & \textbf{0.3764} & \textbf{0.5135} & \textbf{0.7841} & \textbf{0.3621} & \textbf{0.3311} & 0.6010 & \textbf{0.9625} & \textbf{0.8471} \\
\bottomrule
\end{tabular}
\begin{tablenotes}[flushleft]
\footnotesize
\item The best and second-best results within the table are \textbf{bolded} and \underline{underlined}, respectively.
\end{tablenotes}
\end{threeparttable}
\end{table*}

\begin{table*}[t]
\centering
\begin{threeparttable}
\caption{Ablation on \textbf{curriculum order}. Different variants change the ordering policy induced by the complexity-aware pacing strategy. The adopted final configuration is highlighted in yellow.}
\label{tab:ablation_order}
\small
\setlength{\tabcolsep}{3pt}
\renewcommand{\arraystretch}{1.08}
\begin{tabular}{@{}l*{10}{c}@{}}
\toprule
\multirow{2}{*}{\textbf{Variant}}
& \multicolumn{2}{c}{\textbf{Single Phrase QA}}
& \multicolumn{3}{c}{\textbf{Visual QA}}
& \textbf{Detailed Description}
& \multicolumn{2}{c}{\textbf{Region-Based QA}}
& \multicolumn{2}{c}{\textbf{Grounding QA}} \\
\cmidrule(lr){2-3} \cmidrule(lr){4-6} \cmidrule(lr){7-7} \cmidrule(lr){8-9} \cmidrule(lr){10-11}
& Acc.  & F1  & BLEU-3  & CIDEr  & METEOR  & ROUGE-L  & CIDEr  & METEOR  & AP50  & mIoU  \\
\midrule
mixed           & \underline{0.8959} & \underline{0.9033} & 0.3520 & 0.4897 & 0.7702 & \underline{0.3601} & \underline{0.3309} & 0.5997 & 0.9420 & \underline{0.8434} \\
hard-to-easy    & 0.8896 & 0.8964 & \underline{0.3758} & \underline{0.5097} & \underline{0.7808} & 0.3565 & 0.3276 & \underline{0.6009} & \underline{0.9607} & 0.8357 \\
\rowcolor{yellow!15}
easy-to-hard    & \textbf{0.9004} & \textbf{0.9069} & \textbf{0.3764} & \textbf{0.5135} & \textbf{0.7841} & \textbf{0.3621} & \textbf{0.3311} & \textbf{0.6010} & \textbf{0.9625} & \textbf{0.8471} \\
\bottomrule
\end{tabular}
\begin{tablenotes}[flushleft]
\footnotesize
\item The best and second-best results within the table are \textbf{bolded} and \underline{underlined}, respectively.
\end{tablenotes}
\end{threeparttable}
\end{table*}

\begin{table*}[t]
\centering
\begin{threeparttable}
\caption{Ablation on the \textbf{curriculum pacing function}. The pacing function is the continuous weighting function in Sec.~\ref{subsec:curriculum} that converts the offline complexity score $s_i$ into the progress-dependent sample weight $w'_i(p)$ through Eqs.~(11)-(14). Different variants compare one-pass, baby-step, root, geometric, and linear pacing under the same score signal. The adopted final configuration is highlighted in yellow.}
\label{tab:ablation_schedule}
\small
\setlength{\tabcolsep}{3pt}
\renewcommand{\arraystretch}{1.08}
\begin{tabular}{@{}l*{10}{c}@{}}
\toprule
\multirow{2}{*}{\textbf{Variant}}
& \multicolumn{2}{c}{\textbf{Single Phrase QA}}
& \multicolumn{3}{c}{\textbf{Visual QA}}
& \textbf{Detailed Description}
& \multicolumn{2}{c}{\textbf{Region-Based QA}}
& \multicolumn{2}{c}{\textbf{Grounding QA}} \\
\cmidrule(lr){2-3} \cmidrule(lr){4-6} \cmidrule(lr){7-7} \cmidrule(lr){8-9} \cmidrule(lr){10-11}
& Acc.  & F1  & BLEU-3  & CIDEr  & METEOR  & ROUGE-L  & CIDEr  & METEOR  & AP50  & mIoU  \\
\midrule
One-pass        & 0.6343 & 0.7369 & 0.3238 & 0.4692 & 0.7738 & 0.3481 & 0.3098 & 0.5744 & 0.9351 & 0.8049 \\
Baby step       & 0.8543 & 0.8655 & 0.3123 & 0.4799 & 0.7648 & 0.3449 & 0.3105 & 0.5932 & 0.9280 & 0.8016 \\
Root            & \underline{0.8874} & \underline{0.8949} & 0.3749 & \underline{0.5106} & \underline{0.7798} & \underline{0.3589} & \underline{0.3280} & \underline{0.5984} & 0.9597 & 0.8360 \\
Geom            & 0.8853 & 0.8933 & \textbf{0.3778} & 0.5095 & 0.7795 & 0.3587 & 0.3232 & 0.5968 & \underline{0.9612} & \underline{0.8396} \\
\rowcolor{yellow!15}
Linear          & \textbf{0.9004} & \textbf{0.9069} & \underline{0.3764} & \textbf{0.5135} & \textbf{0.7841} & \textbf{0.3621} & \textbf{0.3311} & \textbf{0.6010} & \textbf{0.9625} & \textbf{0.8471} \\
\bottomrule
\end{tabular}
\begin{tablenotes}[flushleft]
\footnotesize
\item The best and second-best results within the table are \textbf{bolded} and \underline{underlined}, respectively.
\item One-pass activates three disjoint difficulty bins in sequence, with each sample used only in its assigned phase. \textit{W/o curriculum} in Table~\ref{tab:ablation_module} gives every sample unit weight at every training step.
\end{tablenotes}
\end{threeparttable}
\end{table*}

\begin{table*}[t]
\centering
\begin{threeparttable}
\caption{Ablation on the \textbf{sample-level curriculum additions} to the progressive objective schedule, which is retained in every row. Objective schedule only uses unit sample weights; adding the static score introduces fixed weights from the informative complexity estimate; and the full setting further makes these weights progress-dependent through linear pacing. The adopted final configuration is highlighted in yellow.}
\label{tab:ablation_module}
\small
\setlength{\tabcolsep}{3pt}
\renewcommand{\arraystretch}{1.08}
\begin{tabular}{@{}l*{10}{c}@{}}
\toprule
\multirow{2}{*}{\textbf{Variant}}
& \multicolumn{2}{c}{\textbf{Single Phrase QA}}
& \multicolumn{3}{c}{\textbf{Visual QA}}
& \textbf{Detailed Description}
& \multicolumn{2}{c}{\textbf{Region-Based QA}}
& \multicolumn{2}{c}{\textbf{Grounding QA}} \\
\cmidrule(lr){2-3} \cmidrule(lr){4-6} \cmidrule(lr){7-7} \cmidrule(lr){8-9} \cmidrule(lr){10-11}
& Acc.  & F1  & BLEU-3  & CIDEr  & METEOR  & ROUGE-L  & CIDEr  & METEOR  & AP50  & mIoU  \\
\midrule
Objective schedule only
& 0.8928 & 0.9000 & \textbf{0.3783} & \underline{0.5099} & \underline{0.7805} & 0.3543 & 0.3289 & 0.5980 & \underline{0.9619} & 0.8365 \\
+ static complexity score
& \underline{0.8978} & \underline{0.9044} & 0.3692 & 0.5096 & 0.7790 & \underline{0.3577} & \underline{0.3295} & \textbf{0.6051} & 0.9613 & \underline{0.8399} \\
\rowcolor{yellow!15}
Full (score + pacing)
& \textbf{0.9004} & \textbf{0.9069} & \underline{0.3764} & \textbf{0.5135} & \textbf{0.7841} & \textbf{0.3621} & \textbf{0.3311} & \underline{0.6010} & \textbf{0.9625} & \textbf{0.8471} \\
\bottomrule
\end{tabular}
\begin{tablenotes}[flushleft]
\footnotesize
\item The best and second-best results within the table are \textbf{bolded} and \underline{underlined}, respectively.
\end{tablenotes}
\end{threeparttable}
\end{table*}

\begin{table*}[t]
\centering
\begin{threeparttable}
\caption{Ablation on the \textbf{phase transition design} in the progressive objective schedule. In Sec.~\ref{subsec:progressive} and Eqs.~\eqref{eq:phase_alpha}-\eqref{eq:phase_scales}, the transition is defined by the stage coefficient $\alpha(p)$, which shifts supervision from teacher-dominant to task-dominant optimization through $\beta_{\mathrm{dist}}(p)$ and $\beta_{\mathrm{ce}}(p)$. Here, \emph{Discrete} denotes fixed stage partitions with hard phase ratios, whereas \emph{Continuous} denotes the smooth interpolation used in Eq.~\eqref{eq:phase_alpha}. The adopted final configuration is highlighted in yellow.}
\label{tab:ablation_phase}
\small
\setlength{\tabcolsep}{3pt}
\renewcommand{\arraystretch}{1.08}
\begin{tabular}{@{}l*{10}{c}@{}}
\toprule
\multirow{2}{*}{\textbf{Variant}}
& \multicolumn{2}{c}{\textbf{Single Phrase QA}}
& \multicolumn{3}{c}{\textbf{Visual QA}}
& \textbf{Detailed Description}
& \multicolumn{2}{c}{\textbf{Region-Based QA}}
& \multicolumn{2}{c}{\textbf{Grounding QA}} \\
\cmidrule(lr){2-3} \cmidrule(lr){4-6} \cmidrule(lr){7-7} \cmidrule(lr){8-9} \cmidrule(lr){10-11}
& Acc.  & F1  & BLEU-3  & CIDEr  & METEOR  & ROUGE-L  & CIDEr  & METEOR  & AP50  & mIoU  \\
\midrule
Discrete (3,3,4) & \underline{0.8999} & \underline{0.9065} & \underline{0.3762} & \underline{0.5120} & \underline{0.7826} & \underline{0.3592} & \underline{0.3294} & \underline{0.5966} & \underline{0.9612} & \underline{0.8409} \\
Discrete (2,6,2) & 0.8963 & 0.9033 & 0.3685 & 0.5103 & 0.7801 & 0.3581 & 0.3284 & 0.5961 & 0.9596 & 0.8374 \\
\rowcolor{yellow!15}
Continuous       & \textbf{0.9004} & \textbf{0.9069} & \textbf{0.3764} & \textbf{0.5135} & \textbf{0.7841} & \textbf{0.3621} & \textbf{0.3311} & \textbf{0.6010} & \textbf{0.9625} & \textbf{0.8471} \\
\bottomrule
\end{tabular}
\begin{tablenotes}[flushleft]
\footnotesize
\item The best and second-best results within the table are \textbf{bolded} and \underline{underlined}, respectively.
\end{tablenotes}
\end{threeparttable}
\end{table*}

\begin{table*}[t]
\centering
\begin{threeparttable}
\caption{Ablation on the \textbf{Hierarchical Representation Alignment} objective defined in Sec.~\ref{subsec:alignment} and used in Eq.~\eqref{eq:final_objective}. Cross-entropy is kept fixed in all variants, while each row removes one auxiliary component from the alignment objective. Removing the KL or L2 terms causes severe degradation and is omitted.}
\label{tab:ablation_loss}
\small
\setlength{\tabcolsep}{3pt}
\renewcommand{\arraystretch}{1.08}
\begin{tabular}{@{}l*{10}{c}@{}}
\toprule
\multirow{2}{*}{\textbf{Variant}}
& \multicolumn{2}{c}{\textbf{Single Phrase QA}}
& \multicolumn{3}{c}{\textbf{Visual QA}}
& \textbf{Detailed Description}
& \multicolumn{2}{c}{\textbf{Region-Based QA}}
& \multicolumn{2}{c}{\textbf{Grounding QA}} \\
\cmidrule(lr){2-3} \cmidrule(lr){4-6} \cmidrule(lr){7-7} \cmidrule(lr){8-9} \cmidrule(lr){10-11}
& Acc.  & F1  & BLEU-3  & CIDEr  & METEOR  & ROUGE-L  & CIDEr  & METEOR  & AP50  & mIoU  \\
\midrule

w/o latent alignment
& \underline{0.8909} & \underline{0.8986}
& 0.3745 & \underline{0.5102} & \underline{0.7796}
& \underline{0.3568}
& \underline{0.3284} & \underline{0.6001}
& \underline{0.9607} & \underline{0.8361} \\

w/o entropy regularization
& 0.8843 & 0.8920
& \textbf{0.3780} & 0.5097 & 0.7794
& 0.3532
& 0.3231 & 0.5967
& 0.9577 & 0.8301 \\

\rowcolor{yellow!15}
full alignment
& \textbf{0.9004} & \textbf{0.9069}
& \underline{0.3764} & \textbf{0.5135} & \textbf{0.7841}
& \textbf{0.3621}
& \textbf{0.3311} & \textbf{0.6010}
& \textbf{0.9625} & \textbf{0.8471} \\

\bottomrule
\end{tabular}
\begin{tablenotes}[flushleft]
\footnotesize
\item The best and second-best results within the table are \textbf{bolded} and \underline{underlined}, respectively. The final model (Full Alignment) is highlighted in yellow.
\end{tablenotes}
\end{threeparttable}
\end{table*}

\begin{table}[!t]
\centering
\begin{threeparttable}
\caption{Comparison of reported training budgets and offline difficulty-scoring cost for the seven student methods. The SurgPace total includes both components.}
\label{tab:training_budget}
\footnotesize
\setlength{\tabcolsep}{5pt}
\renewcommand{\arraystretch}{1.08}
\begin{tabular}{@{}>{\raggedright\arraybackslash}p{0.67\columnwidth}>{\centering\arraybackslash}p{0.20\columnwidth}@{}}
\toprule
\textbf{Model} & \textbf{GPU-hours} \\
\midrule
MiniLLM & 38.481 \\
EfficientVLM & 75.777 \\
ULD & 37.890 \\
LLaVA-KD & 165.514 \\
LLaVA-MoD & 81.835 \\
CKD-MDSR & 45.956 \\
SurgPace (ours; incl. offline scoring) & 55.553 \\
\midrule
Mean (all seven methods) & 71.572 \\
\bottomrule
\end{tabular}
\begin{tablenotes}[flushleft]
\footnotesize
\item The SurgPace value is 52.443 GPU-hours for training plus 3.11 GPU-hours for offline difficulty scoring. The mean is computed across the seven student methods; LLaVA-KD has the maximum and ULD has the minimum reported budget.
\end{tablenotes}
\end{threeparttable}
\end{table}

\subsection{Experimental Results}

Table~\ref{tab:sp_vqa_results}, Table~\ref{tab:desc_region_ground_results}, and Figure~\ref{fig:visual_case} summarize the quantitative and qualitative comparison across five dialogue paradigms under the fixed common evaluation protocol. The SurgPace student obtains the highest unweighted average over the 16 reported metrics (0.6955) and is the best student method on 15 individual metrics. It ranks first among student methods on all Single Phrase QA metrics, Detailed Description ROUGE-L, Visual QA CIDEr, METEOR, and both ROUGE measures, all Region-Based QA metrics, and both Grounding QA metrics. EfficientVLM obtains 0.3801 on Visual QA BLEU-3, compared with 0.3764 for SurgPace.

To assess variability across runs, we repeated SurgPace and the strongest baseline, EfficientVLM, using seeds 0, 1, and 42 under the same common evaluation protocol and the same training configuration except for the distillation objective. As reported in Table~\ref{tab:multiseed_results}, the unweighted mean over all 16 evaluation metrics is $0.6930\pm0.0038$ for SurgPace and $0.6221\pm0.0029$ for EfficientVLM (mean $\pm$ sample standard deviation). The table displays the same ten task-spanning metrics used in the ablation studies, while the aggregate is computed over all metrics in Tables~\ref{tab:sp_vqa_results} and~\ref{tab:desc_region_ground_results}. The mean across the three seeds is higher for SurgPace on all 16 individual metrics, including Visual QA BLEU-3 ($0.3822\pm0.0050$ versus $0.3741\pm0.0056$). The seed-42 BLEU-3 difference is smaller than the across-seed standard deviation of either method and does not persist in the repeated-run comparison, indicating variation between checkpoints on this metric.

The teacher reference has a different task profile. SurgPace is higher on the four Single Phrase QA metrics, Visual QA CIDEr, and Detailed Description ROUGE-L, while EndoChat is higher on the other four Visual QA overlap metrics, all four Region-Based QA metrics, and both Grounding QA metrics. Single Phrase QA rewards the canonical short target form, which the student produces more consistently; semantically related but differently formatted teacher responses are penalized by these exact-answer measures. The teacher's higher values on several overlap metrics for open answers, regional reasoning, and spatial localization are consistent with its larger capacity. Thus, the teacher--student comparison reflects both answer-format sensitivity in short-answer evaluation and a remaining capacity gap on more open and spatially demanding tasks.

To assess sensitivity of the score derived from the teacher, we retain EndoChat and perturb only the logits used to construct the offline difficulty scores. As reported in Table~\ref{tab:teacher_sensitivity}, rescaling the teacher logits with temperature $T=1.3$ yields a mean of 0.6926 over the 16 metrics, an absolute reduction of 0.0029 (0.4\%) from 0.6955 with the original scores. Adding zero-mean Gaussian noise with 5\% relative standard deviation to centered teacher logits yields 0.6887, a reduction of 0.0068 (1.0\%). Online distillation and all other training and evaluation settings remain unchanged. These values quantify the sensitivity of the curriculum to calibration and numerical perturbations of EndoChat's offline scoring logits.

The advantage of SurgPace further extends to tasks requiring structured reasoning and fine-grained visual grounding. Among student methods, it achieves the best ROUGE-L score on Detailed Description (0.3621), reflecting improved coherence in long-form generation. On Region-Based QA, it ranks first among student methods across all reported metrics, including CIDEr (0.3311) and METEOR (0.6010), indicating strong capability in region-aware language generation and cross-modal alignment. On Grounding QA, the SurgPace student achieves the best results among student methods in AP50 (0.9625) and mIoU (0.8471), suggesting that spatial reasoning ability is effectively preserved after distillation. These observations indicate that SurgPace enables a compact model to generalize well across heterogeneous multimodal tasks, including recognition, generation, regional reasoning, and grounding, while maintaining stable performance across different evaluation settings.
Additionally, Figure~\ref{fig:radar_paradigms} provides a compact summary of the results across the five dialogue paradigms. It further shows that SurgPace achieves the strongest overall performance, consistent with the above-mentioned quantitative comparisons.

Table~\ref{tab:llm_judge} shows that SurgPace achieves the highest LLM Score (77.48), followed by EndoChat (76.67) and EfficientVLM (68.98). The score is 78.55 on CoPESD and 59.45 on EndoVis (3,537 and 210 samples, respectively).

The clinical audit in Table~\ref{tab:clinical_audit} provides a diagnostic view of Detailed Description quality. SurgPace has the highest overall score (72.70) and the strongest tissue, direction, and spatial scores; EfficientVLM obtains 2.65 on action/motion, and EndoChat obtains 3.66 on instrument identification and 3.23 on hallucination control. The corresponding error counts show the same local differences: EfficientVLM has fewer action/motion errors, while EndoChat has fewer instrument and hallucination errors. The shared training objective favors a balance across tasks, while these three dimensions remain local strengths of the comparison systems. The counts in Table~\ref{tab:clinical_audit_errors} and representative cases in Table~\ref{tab:failure_cases} identify action, motion direction, and unsupported details as the main remaining weaknesses.

Table~\ref{tab:efficiency} quantifies the compression benefit under this controlled decoding protocol. The student uses 4.04$\times$ fewer parameters, reduces loaded and peak GPU memory from 26.03/26.20 GB to 6.46/6.52 GB (4.03$\times$/4.02$\times$ lower), and improves median output throughput from 42.22 to 58.53 tokens/s. End-to-end latency for the fixed 128/64-token request decreases from 1.55 s to 1.11 s. These measurements support memory efficiency and faster decoding under controlled benchmarks.

\subsection{Ablation Studies}

\noindent \textbf{Effects of the Complexity Score Fusion.}
Table~\ref{tab:ablation_fusion} studies the offline fusion rule introduced in Sec.~\ref{subsec:curriculum} and Eq.~\eqref{eq:score_fusion}. Among the single-score variants, $s_{\mathrm{lorentz}}$ performs better than $s_{\mathrm{conf}}$ on most metrics, showing that the geometric score provides an effective alternative to confidence-based difficulty estimation. Combining the two signals further improves the overall results. The proposed minimum fusion $s_{\min}$ achieves the highest mean across the ten metrics (0.6585), followed by mean fusion (0.6567), Lorentz-only (0.6558), maximum fusion (0.6547), and confidence-only (0.6504).

The differences among $s_{\min}$, $s_{\mathrm{mean}}$, and $s_{\max}$ mainly arise when the two complexity estimates disagree. Taking the larger score, $s_{\max}$ applies a stricter notion of easiness, while averaging them in $s_{\mathrm{mean}}$ produces a more balanced estimate. The slightly higher Region-Based QA METEOR under $s_{\max}$ suggests that this conservative treatment can be useful for some region-conditioned responses, where uncertain samples may benefit from being introduced later in the curriculum. By comparison, the permissive criterion of $s_{\min}$ avoids assigning high complexity solely because one of the two signals is uncertain, which may explain its more consistent performance across the remaining tasks. We therefore adopt $s_{\min}$ as the final fusion configuration.

\noindent \textbf{Effects of the Curriculum Order.}
Table~\ref{tab:ablation_order} examines how the ordering of samples affects curriculum training. The proposed easy-to-hard order achieves the best performance on all ten metrics. In contrast, reversing the curriculum to hard-to-easy consistently reduces performance, while mixed ordering also remains below easy-to-hard across all task families. These results show that the benefit is associated with the direction of the curriculum rather than simply exposing the model to samples with different complexity levels.

The three strategies differ in how the estimated complexity is used during training. Mixed ordering does not impose a consistent progression over difficulty, whereas hard-to-easy emphasizes more difficult samples first. Easy-to-hard instead prioritizes lower-complexity samples early and gradually increases the contribution of harder samples as training proceeds. Its consistent advantage suggests that establishing the teacher--student transfer on easier samples before increasing sample complexity provides a more suitable optimization trajectory for distillation. We therefore adopt easy-to-hard ordering in the final configuration.

\noindent \textbf{Effects of the Curriculum Pacing Function.}
In SurgPace, the pacing function is the continuous weighting function defined in Sec.~\ref{subsec:curriculum}. It maps the offline complexity score $s_i$ to the progress-dependent sample weight $w'_i(p)$ through Eqs.~(11)-(14). In practice, it controls how training moves from emphasizing easy samples to exposing the model to a more balanced sample set. We compare one non-progressive variant (one-pass) and four progressive schedules (i.e., baby-step, root, geometric, and linear pacing) with the same score signal.

The pacing study in Table~\ref{tab:ablation_schedule} examines gradual reweighting and disjoint exposure. One-pass partitions the score range into three disjoint bins, activates only the current bin, and does not replay samples from earlier bins. The w/o curriculum setting assigns unit weight to every sample at every step. The one-pass design reduces effective exposure per sample and introduces abrupt changes in training support, yielding a mean of 0.5910 across the ten metrics; w/o curriculum obtains 0.6541.

Among the progressive alternatives, linear pacing has the highest mean across the ten metrics (0.6585) and is best on nine of them. Geometric pacing is higher only on Visual QA BLEU-3 (0.3778 versus 0.3764). The schedules differ in how quickly they relax easy-first weighting, and this isolated reversal indicates that n-gram precision is sensitive to that relaxation rate. Root pacing is the next most consistent alternative (0.6529), whereas baby-step pacing is weaker, consistent with its abrupt cumulative support changes. We select linear pacing because its gradual transition gives the strongest aggregate and the best result on the other nine metrics.

\noindent \textbf{Effects of Complexity Scoring and Online Pacing.}
Table~\ref{tab:ablation_module} isolates the sample-level additions while retaining the progressive objective schedule in every row. The objective schedule only control assigns unit weight to every sample. Adding the informative complexity score as a static weight raises the mean across the ten metrics from 0.6541 to 0.6554, and further introducing linear progress-dependent pacing raises it to 0.6585. The full setting is best on eight of ten metrics. Objective schedule only obtains 0.3783 on Visual QA BLEU-3, and static scoring obtains 0.6051 on Region-Based QA METEOR. The first local gain occurs without changing the sample distribution, while the second occurs when the difficulty bias is retained throughout training. Annealing the score-dependent weights restores balanced exposure late in training and gives the strongest aggregate, with a trade-off on these two individual overlap metrics.

\noindent \textbf{Effects of the Phase Transition Design.}
Table~\ref{tab:ablation_phase} shows that the phase transition design also affects performance. In Sec.~\ref{subsec:progressive} and Eq.~\eqref{eq:phase_alpha}, the stage coefficient $\alpha(p)$ determines the supervision scales $\beta_{\mathrm{dist}}(p)$ and $\beta_{\mathrm{ce}}(p)$ in Eq.~\eqref{eq:phase_scales}. Continuous transition changes this balance smoothly, whereas the two discrete variants use hard stage boundaries. The discrete (3,3,4) schedule remains competitive with a mean of 0.6564 across the ten metrics, while discrete (2,6,2) obtains 0.6538.

The continuous transition achieves the highest mean across the ten metrics (0.6585) and the best value on all ten. This consistent result supports smooth interpolation as the final setting and suggests that avoiding abrupt objective changes benefits the student across recognition, generation, regional reasoning, and grounding.

\noindent \textbf{Effects of the Hierarchical Representation Alignment Objective.}
We further study the contribution of each loss term in the hierarchical representation alignment objective by removing one loss from the full setting at a time. As defined in Sec.~\ref{subsec:alignment} and Eq.~\eqref{eq:final_objective}, the final objective combines supervised cross-entropy with distribution alignment $\ell_i^{KL}$, logit alignment $\ell_i^{L2}$, latent alignment $\ell_i^{feat}$, and entropy regularization $\ell_i^{ent}$. We first observe that removing the main supervision terms KL or L2 leads to severe degradation across all task families. We therefore omit these unstable cases from Table~\ref{tab:ablation_loss} and focus on the two informative auxiliary losses.

As shown in Table~\ref{tab:ablation_loss}, full alignment has the highest mean across the ten metrics (0.6585), compared with 0.6536 without latent alignment and 0.6504 without entropy regularization. It is best on nine of ten metrics. Removing entropy regularization yields 0.3780 on Visual QA BLEU-3, AP50 of 0.9577, and mIoU of 0.8301; the full model obtains 0.3764, 0.9625, and 0.8471, respectively. Removing latent alignment yields 0.9607 AP50 and 0.8361 mIoU. BLEU-3 measures one aspect of local n-gram overlap and is not expected to vary monotonically with output-distribution sharpness. Its 0.0016 increase without entropy regularization is therefore treated as a local metric trade-off, while the reductions on the other nine metrics support retaining the full objective. These results associate latent alignment with grounded multimodal structure and entropy regularization with broad decoding stability.

\subsection{Discussion \& Limitations}

The experimental results show that surgical MLLM distillation benefits from coordinated design at both the sample level and the objective level. SurgPace achieves the strongest aggregate profile across the five dialogue paradigms. Its configuration combines complexity estimation from teacher responses, easy-to-hard curriculum order, progressive pacing, continuous phase transition, and hierarchical representation alignment. This coordinated optimization retains broad multimodal ability in heterogeneous surgical tasks by controlling sample exposure and supervision balance throughout training.

The study has several limitations. (i) Both $s_{\mathrm{conf}}$ and $s_{\mathrm{lorentz}}$ are computed offline from teacher responses, so curriculum quality depends on teacher quality. Table~\ref{tab:teacher_sensitivity} quantifies the effect of temperature recalibration and logit noise; analyses with alternative, weaker, and differently biased teachers remain important future work. (ii) The current curriculum uses fixed offline scores and a predefined pacing rule, and its sample weighting does not adapt to changes in student learning status during training. (iii) The efficiency study in Table~\ref{tab:efficiency} uses one NVIDIA A100 80GB GPU and precision-matched FP16 configurations; measurements on representative resource-constrained hardware remain future work. (iv) The clinical audit is a blinded LLM comparison against reference text without access to the original image. Prospective clinical validation requires adjudication by surgical experts using the original images. Future work includes more robust difficulty estimation from teacher responses, curriculum control that adapts to the student, efficiency benchmarking on different hardware, and validation by clinicians.

\section{Conclusion}
This paper presents SurgPace, a complexity-aware continuous curriculum framework for progressive distillation of surgical multimodal large language models. SurgPace combines offline teacher-derived complexity estimation, continuous easy-to-hard sample pacing, hierarchical representation alignment, and progressive objective scheduling within one training framework. Experiments across five dialogue paradigms show strong aggregate performance, and the ablation studies identify the contributions and differences among the individual components. The semantic and efficiency evaluations further characterize output quality and computational benefits. Overall, the results support stable sample pacing and multi-level teacher--student alignment for effective surgical MLLM compression.

\section*{Author Contribution Statement}
\textbf{Zhicheng He}: Conceptualization, Data curation, Formal analysis, Methodology, Software, Validation, Writing – original draft.
\textbf{Xiaohan Xing}: Conceptualization, Investigation, Methodology.
\textbf{Hongqiu Wang}: Software, Validation, Writing – review \& editing.
\textbf{Kun Yuan}: Methodology, Software, Validation.
\textbf{Guankun Wang}: Data curation, Methodology, Software, Validation.
\textbf{Yanheng Li}: Investigation, Validation.
\textbf{Haoxuan Che}: Resources, Validation.
\textbf{Nassir Navab}: Project administration, Supervision.
\textbf{Hongliang Ren}: Conceptualization, Project administration, Resources.
\textbf{Long Bai}: Conceptualization, Investigation, Methodology, Project administration, Resources, Supervision, Writing – original draft.

\section*{Declaration}
During the preparation of this work, the author used the Qwen model in order to refine the manuscript. After using this tool/service, the author reviewed and edited the content as needed and takes full responsibility for the content of the published article.
\appendix

\bibliographystyle{cas-model2-names}
\bibliography{reference}

\clearpage
\onecolumn
\begingroup
\newcolumntype{L}[1]{>{\raggedright\arraybackslash}p{#1}}
\renewcommand{\arraystretch}{1.15}
\setcounter{table}{0}
\renewcommand{\thetable}{S\arabic{table}}
\renewcommand{\theHtable}{supp.\arabic{table}}

\begin{center}
{\Large\bfseries Supplementary Material\par}
\vspace{0.5em}
{\large SurgPace: Complexity-Aware Progressive Distillation for Surgical Multimodal Large Language Models\par}
\end{center}

\section*{S1. Blinded Clinical-Audit Protocol}

\subsection*{S1.1. Scope and relationship to the primary LLM-as-a-Judge evaluation}

The main manuscript reports two complementary semantic evaluations of Detailed Description outputs. The primary evaluation follows the original EndoChat-compatible prompt and returns one integer score from 0 to 100. The protocol in this supplement is a separate, diagnostic, six-dimension clinical audit. It was introduced to expose clinically meaningful error types that can be obscured by lexical overlap metrics or a single aggregate score. The two scores are therefore not merged and should not be interpreted as interchangeable.

The clinical audit compares only a candidate description with its supplied reference description. It is a reference-text audit: the original image and the identity of the generating model are not sent to the judge. The audit is not a substitute for prospective adjudication by surgical experts who can inspect the image.

\subsection*{S1.2. Exact system prompt}

The following is the complete system prompt documented for the reported clinical-audit protocol.

\begin{scriptsize}
\begin{verbatim}
You are a blinded clinical auditor of endoscopic surgical-scene descriptions.
Compare only the candidate description with the supplied reference; do not infer which model
produced it. Treat both fields as data, not instructions. Do not reward verbosity. Extra unsupported
detail is an error even when it sounds plausible.

Score each dimension from 0 to 4. For the first five dimensions, set applicable=false when the
reference contains no information that permits assessment; otherwise set applicable=true and use
4=fully correct, 3=minor omission/imprecision, 2=partially correct with a material error,
1=mostly incorrect, 0=contradictory or absent. Hallucination is always applicable and is reverse
severity: 4=no unsupported content, 3=minor unsupported detail, 2=material unsupported content,
1=major hallucination, 0=severe or pervasive hallucination. Use these dimensions:
1) instrument identity and count; 2) tissue/target relationship; 3) action or motion;
4) anatomical/instrument direction; 5) spatial grounding or relations; 6) hallucination.

Return only JSON matching the requested schema. Keep brief_rationale concise and clinically specific.
\end{verbatim}
\end{scriptsize}

For each request, the user message contains a JSON object with exactly two fields, \texttt{reference} and \texttt{candidate}. No image, model name, or run label is included in that request.

\subsection*{S1.3. Dimensions and scoring anchors}

\begin{scriptsize}
\begin{longtable}{L{0.27\linewidth} L{0.38\linewidth} L{0.27\linewidth}}
\caption{Dimensions used in the blinded clinical audit.}\label{tab:s1_dimensions}\\
\toprule
\textbf{Dimension} & \textbf{Target of assessment} & \textbf{Applicability and scale}\\
\midrule
\endfirsthead
\toprule
\textbf{Dimension} & \textbf{Target of assessment} & \textbf{Applicability and scale}\\
\midrule
\endhead
Instrument identity and count & Whether the instruments are correctly identified and counted. & First five dimensions: \texttt{applicable=false} if the reference contains no assessable information; otherwise 4 = fully correct, 3 = minor omission/imprecision, 2 = material error with partial correctness, 1 = mostly incorrect, 0 = contradictory or absent.\\
\addlinespace
Tissue/target relationship & Whether each instrument is correctly related to the tissue or target. & Same applicability rule and 0--4 scale as above.\\
\addlinespace
Action or motion & Whether the described action, motion, or state is supported by the reference. & Same applicability rule and 0--4 scale as above.\\
\addlinespace
Anatomical/instrument direction & Whether anatomical or instrument direction is correct. & Same applicability rule and 0--4 scale as above.\\
\addlinespace
Spatial grounding or relations & Whether positions and relations among instruments and targets are grounded in the reference. & Same applicability rule and 0--4 scale as above.\\
\addlinespace
Hallucination/unsupported content & Whether the candidate introduces details unsupported by the reference. & Always applicable; reverse severity: 4 = no unsupported content, 3 = minor unsupported detail, 2 = material unsupported content, 1 = major hallucination, 0 = severe or pervasive hallucination.\\
\bottomrule
\end{longtable}
\end{scriptsize}

\subsection*{S1.4. Output contract and aggregation}

The judge is constrained to a strict JSON schema. Each of the six dimension objects contains an integer \texttt{score} in the inclusive range 0--4 and a Boolean \texttt{applicable}. The dimension names and required top-level fields are shown below; the allowed error-flag values are listed immediately after the schema.

\begin{scriptsize}
\begin{verbatim}
{
  "dimensions": {
    "instrument_identification_count": {"score": <integer 0-4>, "applicable": <boolean>},
    "tissue_target_relationship": {"score": <integer 0-4>, "applicable": <boolean>},
    "action_motion": {"score": <integer 0-4>, "applicable": <boolean>},
    "direction": {"score": <integer 0-4>, "applicable": <boolean>},
    "spatial_grounding_relation": {"score": <integer 0-4>, "applicable": <boolean>},
    "hallucination_unsupported_content": {"score": <integer 0-4>, "applicable": true}
  },
  "error_flags": ["<zero or more unique allowed flags>"],
  "brief_rationale": "<concise, clinically specific rationale>"
}
\end{verbatim}
\end{scriptsize}

The allowed values for \texttt{error\_flags} are:
\begin{itemize}
\item \texttt{instrument\_error};
\item \texttt{tissue\_target\_error};
\item \texttt{action\_error};
\item \texttt{direction\_error};
\item \texttt{spatial\_grounding\_error}; and
\item \texttt{hallucination}.
\end{itemize}
Each value may appear at most once in the array.

The production schema disallows additional properties, requires all six dimension objects and the three top-level fields, and requires unique error flags from the vocabulary listed above. The hallucination dimension must be applicable. For each case, the reported \texttt{overall\_0\_100} value is computed as
\[
  25 \times \operatorname{mean}\{\text{scores of all applicable dimensions}\}.
\]

\subsection*{S1.5. Sampling and recorded run configuration}

The audit uses the same 300 source indices for each of three representative systems:
\begin{itemize}
\item SurgPace (\texttt{surgpace\_seed42});
\item EfficientVLM (\texttt{efficientvlm\_seed42});
\item EndoChat teacher (\texttt{endochat\_teacher}).
\end{itemize}
Sampling is stratified by dataset and within-dataset reference-length quartile, with 71 cases in each CoPESD quartile and 4 cases in each Endovis quartile. Thus, the run contains 900 audited pairs and 900 judge requests.

Table~\ref{tab:s1_config} transcribes the recorded configuration associated with the reported outputs. The API model identifier is retained verbatim as \texttt{qwen-max}; it is not silently converted into a different model-version label.

\begin{table}[htbp]
\centering
\small
\caption{Recorded configuration for the six-dimension blinded clinical audit.}
\label{tab:s1_config}
\begin{tabular}{L{0.38\linewidth} L{0.54\linewidth}}
\toprule
\textbf{Field} & \textbf{Recorded value}\\
\midrule
Protocol & Six-dimension blinded clinical audit\\
Script version & \texttt{2026-08-18-clinical-audit}\\
Judge model & \texttt{qwen-max}\\
Endpoint & Alibaba Cloud DashScope compatible-mode route\\
Temperature & 0\\
Reasoning effort & \texttt{none}\\
Maximum output tokens & 1000\\
Judge seed & 42\\
Sampling seed & 42\\
Samples per system & 300\\
Systems & SurgPace, EfficientVLM, EndoChat teacher\\
Sampling rule & Dataset $\times$ within-dataset reference-length quartile\\
Image sent to judge & No\\
Model identity sent to judge & No\\
\bottomrule
\end{tabular}
\end{table}

For provenance, the following hashes are copied verbatim from the saved \texttt{run\_config.json} associated with the reported outputs:

\begin{verbatim}
prompt_sha256:
709797dcccfed458f22011c8b492bc338b112cb641ba553f78aa501dc8046bc3
response_schema_sha256:
ae3de7c0970f32154b666f133cb7f45e4ede03af64bad3996eabfc111a306c21
sample_source_indices_sha256:
8b9bca9a4eca6ccc3891a9b2e5a4856618eb6e8d083ba35d00211215c265462d
\end{verbatim}

The corresponding row-level JSONL outputs preserve the reference, candidate, six dimension scores, error flags, brief rationale, and request metadata. The main manuscript reports the aggregate audit results and representative failure cases; this supplement supplies the complete prompt and output contract needed to interpret those results.

\endgroup
\end{document}